\documentclass{article}

\usepackage[final,nonatbib]{neurips_2020}

\usepackage[utf8]{inputenc} 
\usepackage[T1]{fontenc}    
\usepackage{hyperref}       
\usepackage{url}            
\usepackage{booktabs}       
\usepackage{amsfonts}       
\usepackage{nicefrac}       
\usepackage{microtype}      
\usepackage{graphicx}
\usepackage{subfigure}
\usepackage{amsmath}
\usepackage{algorithmic}
\usepackage[ruled,vlined]{algorithm2e}
\usepackage{multicol}
\usepackage{commath}
\usepackage{color}
\usepackage{bbm}
\usepackage{multirow}
\usepackage{diagbox}
\usepackage{caption}
\usepackage{wrapfig}
\usepackage{makecell}
\usepackage{amsmath}
\usepackage{amsthm}

\newtheorem{theorem}{Theorem}
\newtheorem{proposition}{Proposition}
\newtheorem{definition}{Definition}

\newtheorem{apptheorem}{Theorem}
\newtheorem{appmytheorem}{Theorem}
\numberwithin{appmytheorem}{subsection}
\newtheorem{appproposition}{Proposition}
\numberwithin{appproposition}{subsection}

\numberwithin{appdefinition}{subsection}
\newtheorem{applemma}{Lemma}
\numberwithin{applemma}{subsection}

\title{Towards Scale-Invariant Graph-related \\Problem Solving by Iterative Homogeneous \\Graph Neural Networks}

\author{%
  Hao Tang \\
  Shanghai Jiao Tong University\\
  \texttt{\href{mailto:tanghaosjtu@gmail.com}{tanghaosjtu@gmail.com}} \\
  \And 
  Zhiao Huang \\ 
  UC San Diego \\ 
  \texttt{\href{mailto:z2huang@eng.ucsd.edu}{z2huang@eng.ucsd.edu}}\\ 
  \And 
  Jiayuan Gu \\ 
  UC San Diego \\ 
  \texttt{\href{mailto:jigu@eng.ucsd.edu}{jigu@eng.ucsd.edu}}\\ 
  \And 
  Bao-Liang Lu \\
  Shanghai Jiao Tong University\\ 
  \texttt{\href{mailto:bllu@sjtu.edu.cn}{bllu@sjtu.edu.cn}} \\
  \And 
  Hao Su\\ 
  UC San Diego \\ 
  \texttt{\href{mailto:haosu@eng.ucsd.edu}{haosu@eng.ucsd.edu}}\\
}

\begin{document}

\maketitle

\begin{abstract}
	Current graph neural networks (GNNs) lack generalizability with respect to scales (graph sizes, graph diameters, edge weights, etc..) when solving many graph analysis problems.
	Taking the perspective of synthesizing graph theory programs, we propose several extensions to address the issue. First, inspired by the dependency of the iteration number of common graph theory algorithms on graph size, we learn to terminate the message passing process in GNNs adaptively according to the computation progress. Second, inspired by the fact that many graph theory algorithms are homogeneous with respect to graph weights, we introduce homogeneous transformation layers that are universal homogeneous function approximators, to convert ordinary GNNs to be homogeneous. Experimentally, we show that our GNN can be trained from small-scale graphs but generalize well to large-scale graphs for a number of basic graph theory problems. It also shows generalizability for applications of multi-body physical simulation and image-based navigation problems.
\end{abstract}

\section{Introduction}

Graph, as a powerful data representation, arises in many real-world applications~\cite{schlichtkrull2018modeling,shang2019end,fan2019graph,battaglia2016interaction, pmlr-v80-sanchez-gonzalez18a,LiuLH18emnlp}.
On the other hand, the flexibility of graphs, including the different representations of isomorphic graphs, the unlimited degree distributions~\cite{muchnik2013origins, seshadri2008mobile}, and the \textit{boundless graph scales}~\cite{ying2018graph, 8280512}, also presents many challenges to their analysis.
Recently, Graph Neural Networks (GNNs) have attracted broad attention in solving graph analysis problems.  
They are permutation-invariant/equivariant by design and have shown superior performance on various graph-based applications~\cite{wu2019comprehensive, zhang2018deep, zhou2018graph,  gilmer2017neural, battaglia2018relational}. 

However, investigation into \textit{the generalizability of GNNs with respect to the graph scale} is still limited. Specifically, we are interested in GNNs that can learn from small graphs and perform well on new graphs of arbitrary scales. 
Existing GNNs~\cite{wu2019comprehensive, zhang2018deep, zhou2018graph, battaglia2018relational} are either ineffective or inefficient under this setting. 
In fact, even ignoring the optimization process of network training, the representation power of existing GNNs is yet too limited to achieve graph scale generalizability. There are at least two issues: 1) By using a pre-defined layer number~\cite{li2014mean, zheng2015conditional, tamar2016value}, these GNNs are not able to approximate graph algorithms whose complexity depends on graph size (most graph algorithms in textbooks are of this kind). The reason is easy to see: For most GNNs, each node only uses information of the 1-hop neighborhoods to update features by message passing, and it is impossible for $k$-layer GNNs to send messages between nodes whose distance is larger than $k$. More formally,  Loukas~\cite{Loukas2020What} proves that GNNs, which fall within the message passing framework, lose a significant portion of their power for solving many graph problems when their width and depth are restricted; and 
2) a not-so-obvious observation is that, the range of numbers to be encoded by the internal representation may deviate greatly for graphs of different scales. For example, if we train a GNN to solve the shortest path problem on small graphs of diameter $k$ with weight in the range of $[0, 1]$, the internal representation could only need to build the encoding for the path length within $[0, k]$; but if we test this GNN on a large graph of diameter $K \gg k$ with the same weight range, then it has to use and transform the encoding for $[0, K]$. The performance of classical neural network modules (e.g. the multilayer perceptron in GNNs) are usually highly degraded on those out-of-range inputs. 

\begin{figure}[tp]
	\centering 
	\includegraphics[width=\textwidth]{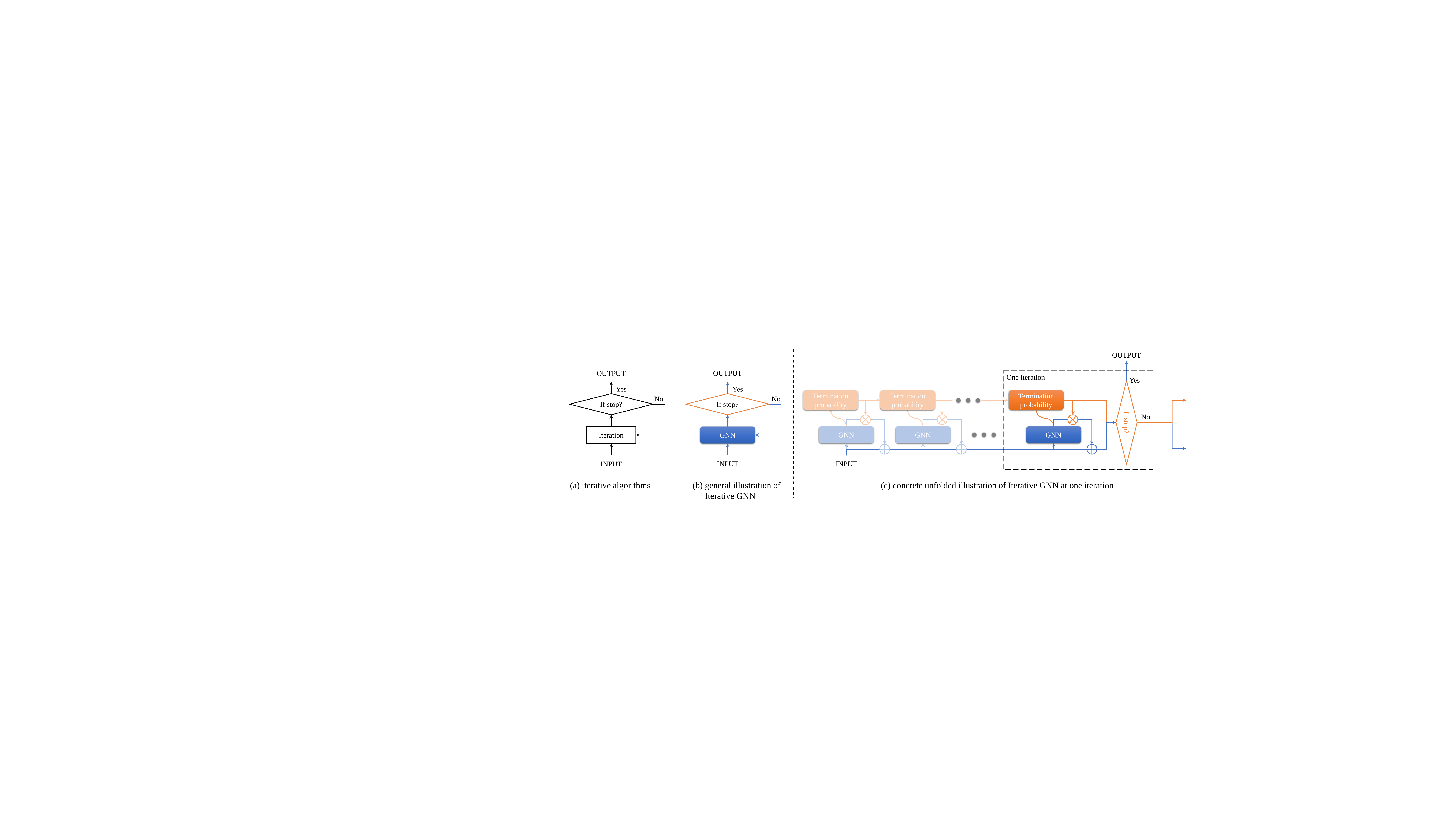}
	\caption{(a) The illustration of general iterative algorithms. The iteration body is repeated until the stopping criterion is satisfied. 
		(b) Illustration of IterGNN as a combination of GNNs and iterative module. 
		(c) A detailed illustration of Iterative GNN. It unfolds the computational flow of IterGNN. Other than the normal data flow (marked as blue), there is another control flow (marked as orange) that serves both as an adaptive stopping criterion and as a data flow controller. 
	}
	\label{fig:itergnn}
\end{figure}

To address the pre-defined layer number issue, we take a program synthesis perspective, to design GNNs that have stronger representation power by mimicking the control flow of classical graph algorithms. Typical graph algorithm, such as Dijkstra's algorithm for shortest path computation, are iterative. They often consist of two sub-modules: an iteration body to solve the sub-problem (e.g., update the distance for the neighborhood of a node as in Dijkstra), and a termination condition to control the loop out of the iteration body. By adjusting the iteration numbers, an iterative algorithm can handle arbitrary large-scale problems. We, therefore, introduce our novel Iterative GNN (IterGNN) that equips ordinary GNN with an adaptive and differentiable stopping criterion to let GNN iterate by itself, as shown in Figure~\ref{fig:itergnn}. 
Our stopping condition is adaptive to the inputs, supports arbitrarily large iteration numbers, and, interestingly, is able to be trained in an end-to-end fashion \emph{without any direct supervision}.

We also give a partial solution to address the issue of out-of-range number encoding, if the underlying graph algorithm is in a specific hypothesis class. More concretely, the solutions to many graph problems, such as the shortest path problem and TSP problem, are homogeneous with respect to the input graph weights, i.e., the solution scales linearly with the magnitudes of the input weights. To build GNNs with representation power to approximate the solution to such graph problems, we further introduce the homogeneous inductive-bias. By assuming the message processing functions are homogeneous, the knowledge that neural networks learn at one scale can be generalized to different scales. We build HomoMLP and HomoGNN as powerful approximates of homogeneous functions over vectors and graphs, respectively.

We summarize our contributions as follows: \textbf{(1)} We propose IterGNN to approximate iterative algorithms, which avoids fixed computation steps in previous graph neural networks, and provides the potential for solving arbitrary large-scale problems. \textbf{(2)} The homogeneous prior is further introduced as a powerful inductive bias for solving many graph-related problems.  \textbf{(3)} We prove the universal approximation theorem of HomoMLP for homogeneous functions and also prove the generalization error bounds of homogeneous neural networks under proper conditions.
\textbf{(4)} In experiments, we demonstrate that our methods can generalize on various tasks and have outperformed baselines.

\section{Related Work}

\textbf{Graph Algorithm Learning.} Despite the success of GNNs (mostly come within the message passing framework~\cite{gilmer2017neural, battaglia2018relational}) in many fields~\cite{zhou2018graph, goyal2018graph, wu2019comprehensive}, few works have reported remarkable results on solving traditional graph-related problems, such as the shortest path problem, by neural networks, especially when the generalizability with regard to scales is taken into account. 
Neural Turing Machine~\cite{graves2014neural, graves2016hybrid} first reported performance on solving the shortest path problem on small graphs using deep neural networks and Neural Logic Machine \cite{dong2018neural} solved the shortest path problem on graphs with limited diameters.
Recently, \cite{velivckovic2019neural}, \cite{velickovic2020pointer} and \cite{xu2019can} achieved positive performance on graph algorithm learning on relatively large graphs using GNNs.   
However, ~\cite{velivckovic2019neural,velickovic2020pointer} require per-layer supervision to train, 
and models in~\cite{xu2019can} can not extend to large graph scales due to their bounded number of message passing steps.
As far as we know, no previous work has solved the shortest path problem by neural networks on graphs of diameters larger than 100. 

\textbf{Iterative Algorithm Approximation.} 
Inspired by the success of traditional iterative algorithms~\cite{dijkstra1959note,blei2017variational}, 
several works were proposed to incorporate the iterative architecture into neural networks for better generalizability~\cite{tamar2016value,selsam2018learning}, more efficiency~\cite{dai2018learning}, or to support end-to-end training~\cite{li2014mean, zheng2015conditional}. However, none of them supports adaptive and unbounded iteration numbers and is therefore not applicable for approximating general iterative algorithms over graphs of any sizes. 

\textbf{Differentiable Controlling Flows.} 
In recent years, multiple works have been proposed in the graph representation learning field that integrate controlling into neural networks to achieve flexible data-driven control. For example, DGCNN~\cite{zhang2018end} implemented a differentiable sort operator (sort pooling) to build more powerful readout functions. Graph U-Net~\cite{gao2019graph, cangea2018towards} designed an adaptive pooling operator (TopK pooling) to support flexible data-driven pooling operations. All these methods achieved the differentiability by relaxing and multiplying the controlling signals with the neural networks' hidden representations. Inspired by their works, our method also differentiates the iterative algorithm by relaxing and multiplying the stopping criterion's output into neural networks' hidden representations. 

\textbf{Adaptive Depth of Neural Networks.} The final formulation of our method is generally similar to the previous adaptive computation time algorithm (ACT)~\cite{graves2016adaptive} for RNNs or spatially ACT~\cite{figurnov2017spatially,eyzaguirre2020differentiable} for CNNs, however, with distinct motivations and formulation details. 
The numbers of iterations for ACT are usually small by design (e.g.the formulation of regularizations and halting distributions). Contrarily,
Our method is designed to fundamentally improve the generalizability of GNNs w.r.t. scales by generalizing to much larger iteration numbers. Several improvements are proposed accordingly. 
The recent flow-based methods
(e.g. the Graph Neural ODE~\cite{poli2019graph})
are also potentially able to provide adaptive layer numbers. However, with no explicit iteration controller, they are not a straightforward solution to approximate iterative algorithms and to encode related inductive biases. 

\section{Backgrounds}
\label{sec:backgrounds}

\textbf{Graphs and graph scales.}
Each graph $G:=(V,E)$ consists of a set of nodes $V$ and a set of edges (pairs of nodes) $E$. To notate graphs with attributes, 
we use $\vec x_v$ for node attributes of node $v\in V$ and use $\vec x_e$ for edge attributes of edge $e\in E$.
We consider three graph properties to quantify the graph scales, which are the number of nodes $N :=|V|$, which is also called the graph size, the graph diameter $\delta_G:=\max_{u,v\in V}d(u,v)$, and the scale of attributes' magnitudes $H:=\max_{v\in V} ||\vec x_v||+\max_{e\in E} ||\vec x_e||$. 
Here, $||\cdot||$ denotes an arbitrary norm of vectors and $d(u,v)$ denotes the length of the shortest path from node $u$ to node $v$, which is also called the distance between node $u$ and node $v$ for undirected graphs.  
We assume graph scales are unbounded but finite, 
and the aim is to generalize learned knowledge to graphs of arbitrary scales.

\textbf{Graph Neural Networks.}
We describe a known class of GNNs that encompasses many state-of-art networks, including GCN~\cite{kipf2017semi},
GAT~\cite{velickovic2018graph}, GIN~\cite{xu2018how}, and Interaction Networks~\cite{battaglia2016interaction}, among others. 
Networks with a global state~\cite{battaglia2018relational} or utilizing multi-hop information per layer~\cite{morris2019weisfeiler, liao2018lanczosnet, isufi2020edgenets} can often be re-expressed within this class, as discussed in~\cite{Loukas2020What}.
The class of GNNs generalizes the message-passing framework~\cite{gilmer2017neural} to handle edge attributes. Each layer of it can be written as 
\begin{eqnarray}
\vec h_v^{(l+1)} & = & f_\theta^{(l)}(\vec h_v^{(l)}, \{\vec x_e:e\in \mathcal{N}_E(v)\}, \{\vec h_{v'}^{(l)}:v'\in \mathcal{N}_V(v)\}).
\label{eq:gnn}
\end{eqnarray}
$\vec h_v^{(l)}$ is the node feature vector of node $v$ at layer $l$. $\mathcal{N}_V(v)$ and $\mathcal{N}_E(v)$ denote the sets of nodes and edges that are directly connected to node $v$ (i.e. its 1-hop neighborhood).
$f_\theta^{(l)}$ is a parameterized  function, 
which is usually composed of  several multilayer perceptron modules and several aggregation functions (e.g. sum/max) in practice.
Readers are referred to~\cite{wu2019comprehensive, zhang2018deep, zhou2018graph, battaglia2018relational} for thorough reviews.

\section{Method}

We propose Iterative GNN (IterGNN) and Homogeneous GNN (HomoGNN) to improve the generalizability of GNNs with respect to graph scales.
IterGNN is first introduced, in Section~\ref{sec:iter-module}, to 
enable adaptive and unbounded iterations of GNN layers so that the model can generalize to graphs of arbitrary scale. 
We further introduce HomoGNN, in Section \ref{sec:homognn}, to partially solve the problem of out-of-range number encoding for graph-related problems.
We finally describe PathGNN that improves the generalizability of GNNs for distance-related problems by improving the algorithm alignments~\cite{xu2019can} to the Bellman-Ford algorithm
in Section \ref{sec:pathgnn}.

\subsection{Iterative module}
\label{sec:iter-module}

\begin{wrapfigure}{R}{0.48\textwidth}
	\begin{minipage}{0.48\textwidth}
		\begin{algorithm}[H]
			\caption{Iterative module. $g$ is the stopping criterion and $f$ is the iteration body}
			\label{alg:iteration}
			\begin{algorithmic}
				\STATE {\bfseries input:} initial feature $x$; stopping threshold $\epsilon$
				\STATE $k\gets 1$
				\STATE $h^0\gets x$
				\WHILE{$\prod_{i=1}^{k-1}(1-c^i)>\epsilon$}
				\STATE $h^k\gets f(h^{k-1})$
				\STATE $c^k\gets g(h^k)$
				\STATE $k\gets k+1$
				\ENDWHILE
				\STATE  {\bfseries return }$h=\sum_{j=1}^{k-1}\left(\prod_{i=1}^{j-1}(1-c^i)\right)c^jh^j$
			\end{algorithmic}
		\end{algorithm}
	\end{minipage}
\end{wrapfigure}

The core of IterGNN is a differentiable iterative module. It executes the same GNN layer repeatedly until a learned stopping criterion is met. 
We present the pseudo-codes in Algorithm~\ref{alg:iteration}.
At time step $k$, the iteration body $f$ updates the hidden states as $h^k=f(h^{k-1})$; the stopping criterion function $g$ then calculates a confidence score $c^k=g(h^k)\in[0,1]$ to describe the probability of the iteration to terminate at this step. The module determines the number of iterations using a random process based on the confidence scores $c^k$. At each time step $k$, the random process has a probability of $c^k$ to terminate the iteration and to return the current hidden states $h^k$ as the output. The probability for the whole process to return $h^k$ is then $p^k=\left(\prod_{i=1}^{k-1}(1-c^i)\right)c^k$, which is the product of the probabilities of continuing the iteration at steps from 1 to $k-1$ and stopping at step $k$. However, the sampling procedure is not differentiable. Instead, we execute the iterative module until the ``continue'' probability $\prod_{i=1}^{k-1}(1-c^i)$ is smaller than a threshold $\epsilon$ and return an expectation $h=\sum_{j=1}^k p^jh^j$ at the end. The gradient to the output $h$ thus can optimize the hidden states $h^k$ and the confidence scores $c^k$ jointly. 

For example, assume $c^i=0$ for $i<k$, $c^{k}=a$, $c^{k+1}=b$, and $(1-a)(1-b)<\epsilon$. If we follow the pre-defined random process, for steps before $k$, the iteration will not stop as $c^i=0$ for $i<k$. For the step $k$, the process has a probability of $a$ to stop and output $h^k$; otherwise, the iteration will continue to the step $k+1$. Similarly, at the step $k+1$, the iteration has a probability of $b$ to stop and output $h^{k+1}$. We stop the iteration after step $k+1$ as the ``continue'' probability $\prod_{i=1}^k (1-c^i)=(1-a)(1-b)$ is negligible.
The final output is the expectation of the output of the random process $h=ah^k+(1-a)bh^{k+1}$.

By setting $f$ and $g$ as GNNs, we obtain our novel IterGNN, as shown in Figure~\ref{fig:itergnn}. The features are associated with nodes in the graph as $\{\vec h_v^{(k)}:v\in V\}$. GNN layers as described in Eq. \ref{eq:gnn} are adopted as the body function $f$ to update the node features iteratively $\{\vec{h}_v^{(k)}:v\in V\}=\text{GNN}(G, \{\vec{h}_v^{(k-1)}:v\in V\}, \{\vec{h}_e:e\in E\})$. We build the termination probability module as $g$ by integrating a readout function and an MLP. The readout function (e.g. max/mean pooling) summarizes all node features $\{\vec{h}_v^{(k)}:v\in V\}$ into a fixed-dimensional vector $\vec h^{(k)}$. The MLP predicts the confidence score as $c^k=\text{sigmoid}(\text{MLP}(\vec h^{(k)}))$. The sigmoid function is utilized to ensure the output of $g$ is between $0$ and $1$. With the help of our iterative module, IterGNN can adaptively adjust the number of iterations. Moreover,  it can be trained without any supervision of the stopping condition. 

Our iterative module can resemble the control flow of many classical graph algorithms since the iteration of most graph algorithms depends on the size of the graph. For example, Dijkstra's algorithm~\cite{dijkstra1959note} has a loop to greedily propagate the shortest path from the source node. The number of iterations to run the loop depends linearly on the graph size. Ideally, we hope that our $f$ can learn the loop body and $g$ can stop the loop when all the nodes have been reached from the source. Interestingly, the experiment result shows such kind behavior. This structural level of the computation allows superior generalizability, which agrees with the findings in~\cite{xu2019can} that improved algorithm alignment can increase network generalizability. 
In contrast, without a dynamic iterative module, previous GNNs have much inferior ability to generalize to larger graphs.

We state more details of IterGNN in Appendix, including the memory-efficient implementation, the theoretical analysis of representation powers, the node-wise iterative module to support unconnected graphs, 
and the decaying confidence mechanism to achieve much larger iteration numbers
during inference in practice
(by compensating the nonzero properties of the sigmoid function in $g$).

\subsection{Homogeneous prior}
\label{sec:homognn}
The homogeneous prior is introduced to improve the generalizability of GNNs for out-of-range
features/attributes. We first define the positive homogeneous property of a function:
\begin{definition}
	\label{def:pos-homo}
	A function $f$ over vectors is positive homogeneous iff $f(\lambda \vec x) = \lambda f(\vec x)$ for all $\lambda>0$.
	
	A function $f$ over graphs is positive homogeneous iff for any graph $G=(V,E)$ with node attributes $\vec{x}_v$ and edge attributes $\vec{x}_e$, $f(G, \{\lambda\vec{x}_v:v\in V\}, \{\lambda\vec{x}_e:e\in E\}) = \lambda f(G, \{\vec{x}_v:v\in V\}, \{\vec{x}_e:e\in E\})$
\end{definition}
The solutions to most graph-related problems are positive homogeneous, such as the length of the shortest path, the maximum flow, graph radius, and the optimal distance in the traveling salesman problem. 

\begin{figure}[tp]
	\centering 
	\includegraphics[width=0.8\textwidth]{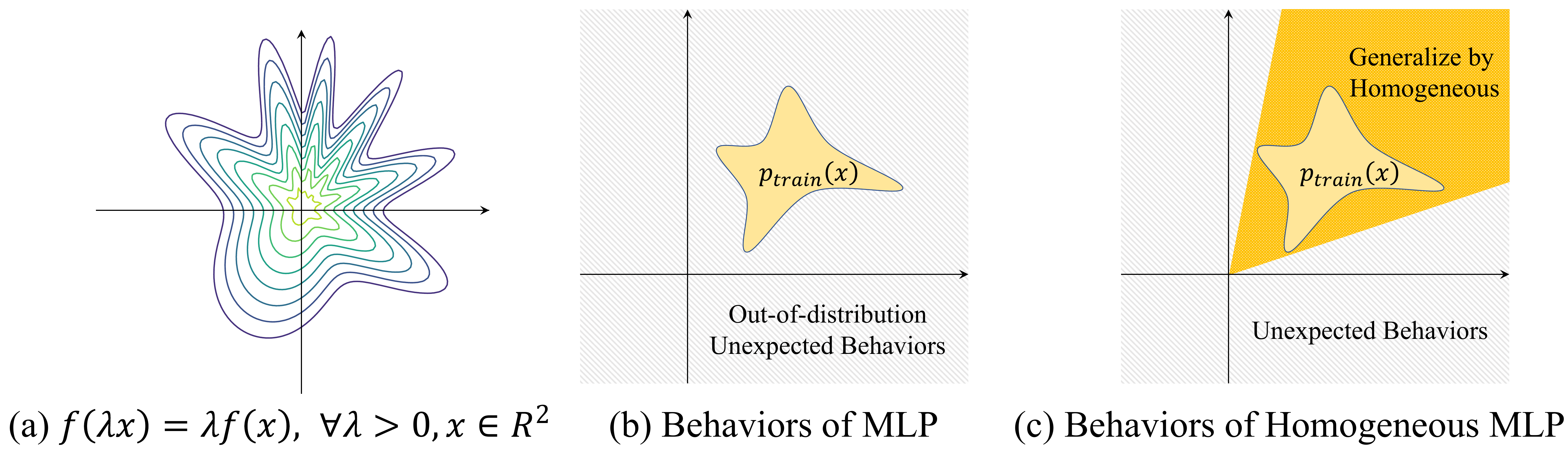}
	\caption{(a) An example of homogeneous functions. 
		(b-c) Illustration of the improved generalizability
		 by applying the homogeneous prior. Knowledge learned from the training samples not only can be generalized to samples of the same data distribution as ordinary neural networks, as shown in (b), but also can be generalized to samples of the scaled data distributions, as shown in (c).
	}
	\label{fig:homognn}
\end{figure}

The homogeneous prior tackles the problem of different magnitudes of features for generalization.
As illustrated in Figure \ref{fig:homognn}, by assuming functions as positive homogeneous, models can generalize knowledge to the scaled features/attributes of different magnitudes. 
For example, let us assume two datasets $D$ and $D_\lambda$ that are only different on magnitudes, which means $D_\lambda:=\{\lambda x: x\in D\}$ and $\lambda > 0$. If the target function $f$ and the function $F_\mathcal{A}$ represented by neural networks $\mathcal{A}$ are both homogeneous, the prediction error on dataset $D_\lambda$ then scales linearly w.r.t. the scaling factor $\lambda$: 
\begin{equation}
\sum_{x\in D_\lambda}||f(x)-F_\mathcal{A}(x)|| = \sum_{x'\in D}||f(\lambda x')-F_\mathcal{A}(\lambda x')|| = \lambda \sum_{x'\in D}||f(x')-F_\mathcal{A}(x')||.
\end{equation}

We design the family of GNNs that are homogeneous, named HomoGNN, as follows:
simply remove all the bias terms in the multi-layer perceptron (MLP) used by ordinary GNNs, so that all affine transformations degenerate to linear transformations. Additionally, only homogeneous activation functions are allowed to be used. Note that ReLU is a homogeneous activation function. The original MLP used in ordinary GNNs become HomoMLP in HomoGNNs afterward.

\subsubsection{Theoretical analysis of HomoGNN and HomoMLP}
\label{sec:homo-theorem}
We provide theoretical proofs showing that, if the target function is homogeneous, low generalization errors and low training errors are both achievable using the pre-defined homogeneous neural networks under proper conditions. 
We first formalize the generalization error bounds of homogeneous neural networks on approximating homogeneous functions under some assumptions, by extending the previous example to more general cases. 
To show that low training errors are achievable, we further 
prove that HomoMLP is a universal approximator of the homogeneous functions under proper conditions, based on the universal approximation theorem for width-bounded ReLU networks~\cite{NIPS2017_7203}. We present propositions stating that HomoGNN and HomoMLP can only represent homogeneous functions, along with the proofs for all theorems, in the Appendix.

Let training samples $D_m=\{x_1, x_2, \cdots x_m\}$ be independently sampled from the distribution $\mathcal{D}_x$, then if we scale the training samples with the scaling factor $\lambda \in \mathbb{R}^+$ which is independently sampled from the distribution $\mathcal{D}_\lambda$, we get a ``scaled'' distribution $\mathcal{D}_x^\lambda$, which has a  density function $P_{\mathcal{D}_x^\lambda}(z) := \int_{\lambda}\int_{x} \delta(\lambda x=z)P_{\mathcal{D}_\lambda}(\lambda)P_{\mathcal{D}_x}(x)\dif x \dif\lambda$. The following theorem bounds the generalization error bounds on $\mathcal{D}_x^{\lambda}$: 
\begin{theorem} \upshape{(Generalization error bounds of homogeneous neural networks with independent scaling assumption)}. \itshape 
    For any positive homogeneous functions function $f$ and neural network $F_\mathcal{A}$, let $\beta$ bounds the generalization errors on the training distribution $D_x$ , i.e.,  
	$\mathbb{E}_{x\sim\mathcal{D}_x}|f(x)-\mathcal{F}_\mathcal{A}(x)| \le \frac{1}{m}\sum_{i=1}^m |f(x_i)-F_\mathcal{A}(x_i)|+\beta$, then the generalization errors on the scaled distributions $\mathcal{D}_x^\lambda$ scale linearly
	with the expectation of scales $\mathbb{E}_{\mathcal{D}_\lambda}[\lambda]$:
	\begin{eqnarray}
	\mathbb{E}_{x\sim\mathcal{D}_x^\lambda}|f(x)-F_\mathcal{A}(x)|= \mathbb{E}_{\mathcal{D}_\lambda}[\lambda] \mathbb{E}_{x\sim\mathcal{D}_x} |f(x)-F_\mathcal{A}(x)| \le \mathbb{E}_{\mathcal{D}_\lambda}[\lambda](\frac{1}{m}\sum_{i=1}^m |f(x_i)-F_\mathcal{A}(x_i)|+\beta)
	\end{eqnarray}
	\label{thm: homo-generalize}
\end{theorem}
\begin{theorem} \upshape {(Universal approximation theorem for width-bounded HomoMLP).} \itshape
	For any positive-homogeneous Lebesgue-integrable function $f:\mathbb{X}\mapsto \mathbb{R}$, where $\mathbb{X}$ is a Lebesgue-measurable compact subset of $\mathbb{R}^n$, and for any $\epsilon>0$, there exists a finite-layer HomoMLP $\mathcal{A}'$ with width $d_m\le 2(n+4)$, which represents the function $F_{\mathcal{A}'}$ such that 
	$\int_{\mathbb{X}}|f(x)-F_{\mathcal{A}'}(x)|\dif x < \epsilon$.
\end{theorem}

\subsection{Path graph neural networks}
\label{sec:pathgnn}

We design PathGNN to imitate one iteration of the classical Bellman-Ford algorithm.
It inherits the generalizability of the Bellman-Ford algorithm and the flexibility of the neural networks. 
Specifically, the Bellman-Ford algorithm performs the operation $dist_i=\min(dist_i,\min_{j\in\mathcal{N}(i)}(dist_j+w_{ji}))$ iteratively to solve the shortest path problem, where $dist_i$ is the current estimated distance from the source node to the node $i$, and $w_{ji}$ denotes the weight of the edge from node $j$ to node $i$. If we consider $dist_i$ as node features and $w_{ij}$ as edge features, one iteration of the Bellman-Ford algorithm can be exactly reproduced by GNN layers as described in Eq. \ref{eq:gnn}:
\begin{equation*}
\vec h_i=\min(\vec h_i,\min_{j\in\mathcal{N}(i)}(\vec h_j+\vec x_{ji}))\equiv -\max(-\vec h_i,\max_{j\in\mathcal{N}(i)}(-\vec h_j-\vec x_{ji})).
\end{equation*}
To achieve more flexibilities for solving problems other than the shortest path problem, 
we integrate neural network modules, such as MLPs to update features or the classical attentional-pooling to aggregate features, while building the PathGNN layers. A typical variant of PathGNN is as follows:
\begin{eqnarray*}
& \alpha_{ji} = \text{softmax}(\{\text{MLP}_1(\vec h_j;\vec h_i;\vec x_{ji})~\text{for}~j \in \mathcal{N}(i)\}); &\\
& \vec h'_i = \sum_{j\in\mathcal{N}(i)}\alpha_{ji}\text{MLP}_2(\vec h_j;\vec h_i;\vec x_{ji});
~~~~\vec h_i~=~\max(\vec h_i,\vec h'_i). &
\end{eqnarray*}
We state the detailed formulation and variations of PathGNN layers in the Appendix.

\section{Experiments}

Our experimental evaluation aims to study the following empirical questions: \textbf{(1)} Will our proposals, the PathGNN layer, the homogeneous prior, and the iterative module, improve the generalizability of GNNs with respect to graph scales that are the number of nodes, the diameter of graphs, and the magnitude of attributes? \textbf{(2)} Will our iterative module adaptively change the iteration numbers and consequently learn an interpretable stopping criterion in practice? \textbf{(3)} Can our proposals improve the performance of general graph-based reasoning tasks such as those in physical simulation, image-based navigation, and reinforcement learning? 

\begin{figure}
	\centering
	\subfigure[Physical Simulation]{\includegraphics[width=0.31\textwidth]{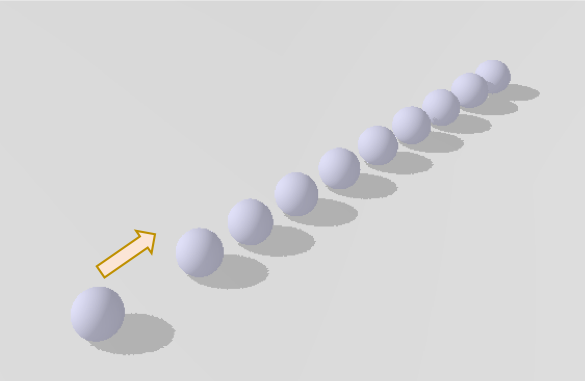} \label{fig:newton}}%
	\qquad
	\subfigure[Symbolic PacMan]{\includegraphics[width=0.2\textwidth]{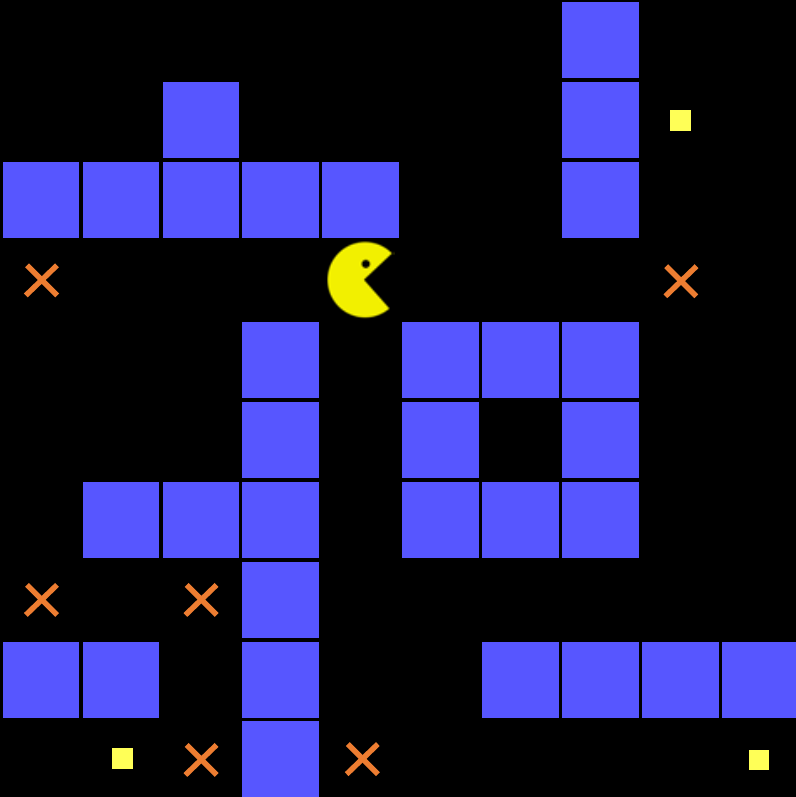}\label{fig:symbolic_pacman}}%
	\qquad
	\subfigure[Image-based Navigation]{\includegraphics[width=0.31\textwidth]{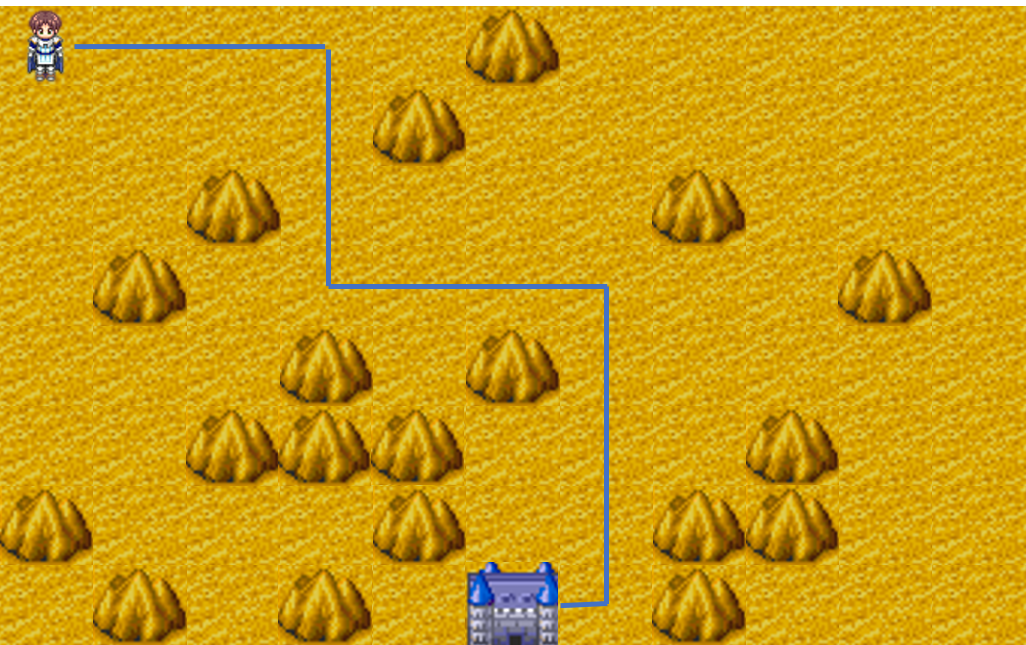}\label{fig:image-navi}}
	\caption{
		Figure (a) shows a set of Newton's balls in the physical simulator. The yellow arrow is the moving direction of the first ball. 
		Figure (b) shows our symbolic PacMan environment. Figure (c) illustrates our image-based navigation task in a RPG-game environment.}
	\label{fig:general-task}
\end{figure}{}

\textbf{Graph theory problems and tasks.}
We consider three graph theory problems, i.e., shortest path, component counting, and Traveling Salesman Problem (TSP), to evaluate models' generalizability w.r.t. graph scales.
We build a benchmark by combining multiple graph generators, including  Erdos-Renyi (ER), K-Nearest-Neighborhoods graphs (KNN), planar graphs (PL), and lobster graphs (Lob), so that the generated graphs can have more diverse properties. 
We further apply our proposals to three graph-related reasoning tasks, i.e., physical simulation, symbolic Pacman, and image-based navigation, as illustrated in Figure \ref{fig:general-task}. 
The generation processes and the properties of datasets are listed in the Appendix.

\textbf{Models and baselines.}
Previous problems and tasks can be formulated as graph regression/classification problems. We thus construct models and baselines following the common practice~\cite{battaglia2018relational,zhang2018end,xu2018how}.
We stack 30 GCN~\cite{kipf2017semi}/GAT~\cite{velickovic2018graph} layers to build the baseline models. GIN~\cite{xu2018how} is not enlisted since 30-layer GINs do not converge in most of our preliminary experiments. Our ``Path'' model stacks 30 PathGNN layers. Our ``Homo-Path'' model replaces GNNs and MLPs in the ``Path'' model with HomoGNNs and HomoMLPs.
Our ``Iter-Path'' model adopts the iterative module to control the iteration number of the GNN layer in the ``Path'' model. 
The final ``Iter-Homo-Path'' integrates all proposals together.
Details are in the Appendix.

\textbf{Training Details.}
We utilize the default hyper-parameters to train models. We generate 10000 samples for training, 1000 samples for validation, and 1000 samples for testing. The only two tunable hyper-parameter in our experiment is the epoch number (10 choices) and the formulation of PathGNN layers (3 choices). Validation datasets are used to tune them. Specially for the component counting problem, we incorporate the random initialization of node features so that GNNs have more representation powers to distinguish non-isomorphic graphs~\cite{xu2018how, morris2019weisfeiler,murphy2019relational, anonymous2020coloring}. More details are listed in the Appendix.

\subsection{Solving graph theory problems}
\label{sec:exp.graph-theory}

\textbf{Generalize w.r.t. graph sizes and graph diameters.}
\label{sec:exp.gen-size-diameter}
We present the generalization performance for all three graph theory problems in Table \ref{tab:summary}. Models are trained on graphs of sizes within $[4,34)$ and are evaluated on graphs of larger sizes such as 100 (for shortest path and TSP) and 500 (for component counting so that the diameters of components are large enough). 
The relative loss metric is defined as $|y-\hat y|/|y|$, given a label $y$ and a prediction $\hat y$.
The results demonstrate that each of our proposals improves the generalizability on almost all problems. Exceptions happen on graphs generated by ER.
It is because the diameters of those graphs are 2 with high probability even though the graph sizes are large. 
Our final model, \mbox{Iter-Homo-Path}, which integrates all proposals, performs much better than the baselines such as GCN and GAT. 
The performance on graphs generated by KNN and PL further supports the analysis. The concrete results are presented in the Appendix due to space limitations. 
We also evaluated a deeper Path model, i.e., with 100 layers, on the weighted shortest path problem (Lob). The generalization performance (relative loss$\approx$ 0.13) became even worse. 

We then explore models' generalizability on much larger graphs on the shortest path problem
using Lob to generate graphs with larger diameters.
As shown in Table \ref{tab:generalize-on-lobster}, our model achieves a \mbox{100\% success} rate of identifying the shortest paths on graphs with as large as 5000 nodes even though it is trained on graphs of sizes within $[4,34)$.
As claimed, the iterative module is necessary for generalizing to graphs of much larger sizes and diameters due to the message passing nature of GNNs. The iterative module successfully improves the performance from $\sim 60\%$ to 100\% on graphs of sizes $\ge500$. 

\textbf{Ablation studies and comparison.}
We conduct ablation studies to exhibit the benefits of our proposals using the unweighted shortest path problem on lobster graphs with 1000 nodes in Table~\ref{tab:shortest_path_ablation_500}. The models are built by replacing each proposal in our best Iter-Homo-Path model with other possible substitutes in the literature. For the iterative module, other than the simplest paradigm utilized in \mbox{Homo-Path} that stacks GNN layers sequentially, we also compare it with the ACT algorithm~\cite{graves2016adaptive} and the fixed-depth weight-sharing paradigm~\cite{tamar2016value,li2014mean}, resulting in the ``ACT-Homo-Path'' and ``Shared-Homo-Path'' models. The ACT algorithm provides adaptive but usually short iterations of layers 
(see Figure \ref{fig:iteration-number} and Appendix).
The weight-sharing paradigm iterates modules for predefined times and assumes that the predefined iteration number is large enough. We set its iteration number to the largest graph size in the dataset.
Homo-Path and ACT-Homo-Path perform much worse than Iter/Shared-Homo-Path because of the limited representation powers of shallow GNNs. 
Shared-Homo-Path performs worse than our Iter-Homo-Path, possibly because of the accumulated errors after unnecessary iterations.
For the homogeneous prior, we build ``Iter-Path'' by simply removing the homogeneous prior. It performs much worse than Iter-Homo-Path because of the poor performance of MLPs on out-of-distribution features. 
For PathGNN, we build ``Iter-Homo-GCN'' and ``Iter-Homo-GAT'' by replacing PathGNN with GCN and GAT. Their bad performance verifies the benefits of better algorithm alignments~\cite{xu2019can}.

\begin{table}
	\caption{
		Generalization performance on graph algorithm learning and graph-related reasoning. 
		Models are trained on graphs of smaller sizes (e.g., within $[4,34)$ or $\le 10\times 10$) and are tested on graphs of larger sizes (e.g., 50, 100, 500, $16\times16$ or $33\times 33$).
		The metric for the shortest path and TSP is the relative loss. The metric for component counting is accuracy. The metric for physical simulation is the mean square error. The metric for image-based navigation is the success rate.
	}
\label{tab:summary}
\small
\centering
	\begin{tabular}{|c||cc|cc|c||cc|cc|}
	\hline
	& \multicolumn{5}{c||}{Graph Theory Problems} &  \multicolumn{4}{c|}{Graph-related Reasoning}
	\\ \cline{2-10}
	& \multicolumn{2}{c|}{Shortest Path} & \multicolumn{2}{c|}{Component Cnt.} &  \multicolumn{1}{c||}{TSP}
	 &  \multicolumn{2}{c|}{Physical sim.} &  \multicolumn{2}{c|}{Image-based Navi.}\\
	Models & ER & Lob & ER & Lob & 2D & 50 & 100 & $16\times 16$ & $33\times 33$ 
	\\\hline
	GCN~\cite{kipf2017semi} & 0.1937 & 0.44& 0.0\% & 0.0\%& 0.52 & 42.18 & 121.14 & 34.2\% & 28.9\% 
	\\
	GAT~\cite{velickovic2018graph} & 0.1731 & 0.28 & 24.4 \% & 0.0\%& 0.18 & >1e4 &  >1e4 & 56.7\% & 44.5\% \\
	Path (ours) & \textbf{0.0003} & 0.29 & 82.3\% & 77.2\%& 0.16 & 20.24 & 27.67 & 85.6\% & 65.1\% \\
	Homo-Path (ours) & 0.0008 & 0.27 & 91.9\% & 83.9\%& 0.14 & 20.48 & 21.45 & 87.8\% & 69.3\% \\ 
	Iter-Path (ours) & 0.0005 & 0.09 & 86.7\% & 96.1\% & 0.08 & \textbf{0.13} & \textbf{1.68} & 89.4\% & 78.6\% \\
	Iter-Homo-Path (ours) & 0.0007 & \textbf{0.02} & \textbf{99.6\%}& \textbf{97.5\%}& \textbf{0.07} & \textbf{0.07} & \textbf{2.01} & \textbf{98.8\%} & \textbf{91.7\%} \\ \hline
\end{tabular}
\end{table}

\begin{table}
	\centering
	\caption{
		Generalization performance
		on the shortest path problem with lobster graphs. 
		During training, node numbers are within $[4,34)$ for unweighted problems (whose metric is the success rate), and edge weights are within $[0.5,1.5)$ for weighted problems (whose metric is the relative loss).
	}
	\small
	\begin{tabular}{|c||ccccc||cccc|}
		\hline
		\multirow{2}{*}{Generalize} & \multicolumn{5}{c||}{w.r.t.  sizes and diameters - unweighted} & \multicolumn{4}{c|}{w.r.t. magnitudes - weighted}
		\\
		& 20 & 100 & 500 & 1000 & 5000 & $[0.5,1.5)$ & $[1,3)$ & $[2,6)$ & $[8, 24)$
		\\
		\hline
		GCN~\cite{kipf2017semi} & 66.6 & 25.7 & 5.5 & 2.4 & 0.4 & 0.31 & 0.37 & 0.49 & 0.56
		\\
		GAT~\cite{velickovic2018graph} & \textbf{100.0} & 42.7 &  10.5 & 5.3 & 0.9 & 0.13 & 0.29 & 0.49 & 0.55\\
		Path (ours) & \textbf{100.0} & 62.9 & 20.1 & 10.3 & 1.6 & 0.06 & 0.22 & 0.44 & 0.54\\
		Homo-Path (ours) & \textbf{100.0} & 58.3 & 53.7 & 50.2 & 1.6 & 0.03 & \textbf{0.03} & \textbf{0.03} & \textbf{0.03} \\ 
		Iter-Homo-Path (ours) &\textbf{100.0} & \textbf{100.0} & \textbf{100.0} & \textbf{100.0} & \textbf{100.0} & \textbf{0.01} & 0.04 & 0.06 & 0.08\\ 
		\hline
	\end{tabular}
	\label{tab:generalize-on-lobster}
\end{table}

\begin{table}
	\begin{minipage}[b]{0.5\linewidth}
		\centering
		\small
		\begin{tabular}{|c|c|}
			\hline
			\multicolumn{2}{|c|}{Iter-Homo-Path}\\\hline
			\multicolumn{2}{|c|}{\textbf{100.0}}\\
			\hline
			Homo-Path & Iter-Path\\
			\hline
			53.7 & 48.9\\
			\hline
			ACT-Homo-Path & Iter-Homo-GAT\\
			\hline
			52.7 & 2.9 \\
			\hline
			Shared-Homo-Path & Iter-Homo-GCN \\
			\hline
			91.7 & 1.4  \\
			\hline
		\end{tabular}
		\normalsize
		\caption{
			Ablation studies of generalization performance for the shortest path problem on lobster graphs with 1000 nodes. 
			Metric is the success rate.
		}
		\label{tab:shortest_path_ablation_500}
	\end{minipage}\hfill
	\begin{minipage}[b]{0.48\linewidth}
		\centering
		\includegraphics[width=0.95\textwidth]{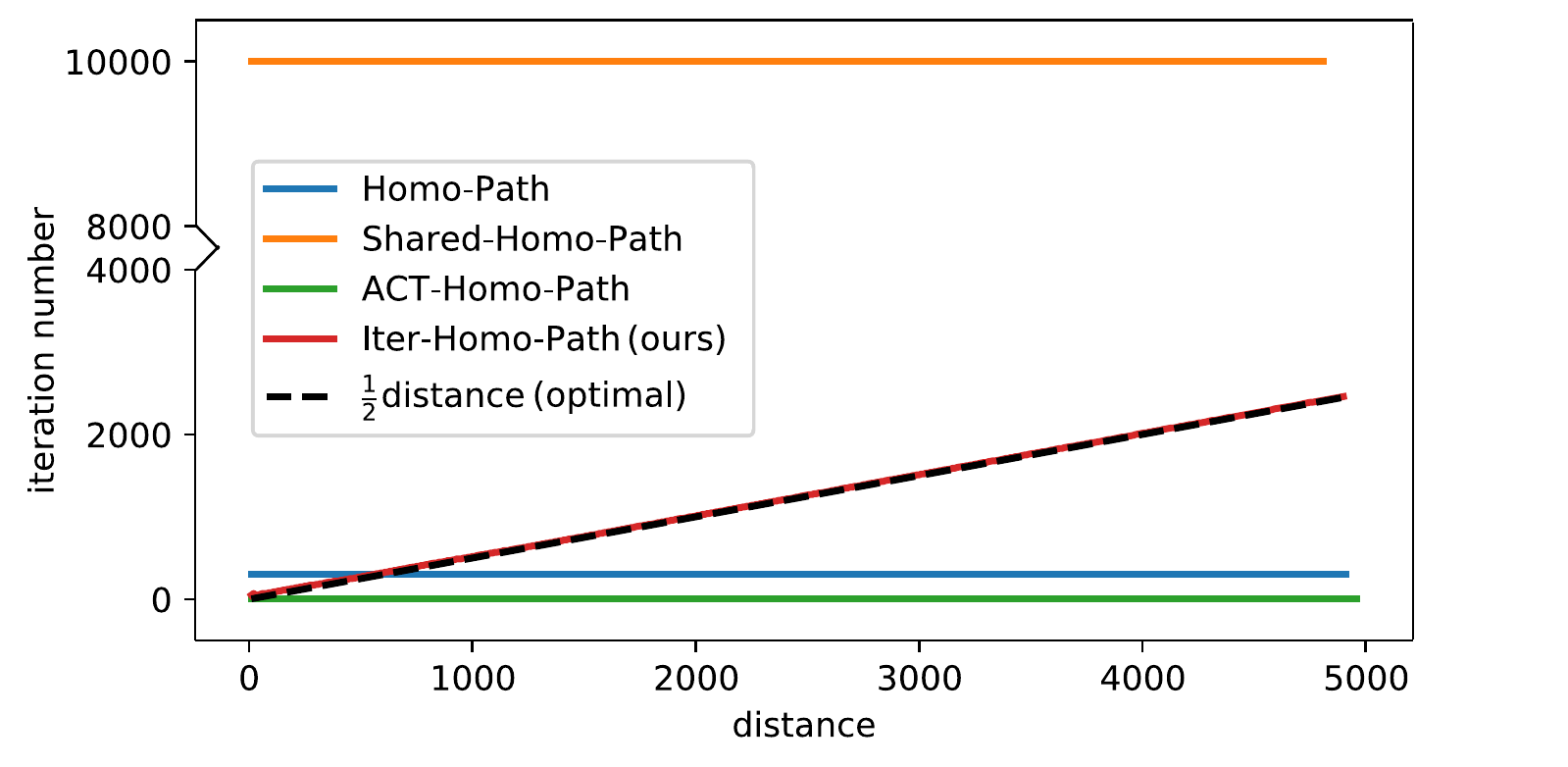}
		\captionof{figure}{The iteration numbers of GNN layers w.r.t. the distances from the source node to the target node for the shortest path problem.}
		\label{fig:iteration-number}
	\end{minipage}
\end{table}

\textbf{Generalize w.r.t. magnitudes of attributes.}
\label{sec:exp.magnitude}
We evaluate the generalizability of models w.r.t. magnitudes of attributes using the weighted shortest path length problem, as shown in Table \ref{tab:generalize-on-lobster}. The edge weights are randomly sampled from $[0.5, 1.5)$ during training and are sampled from $[1,3)$, $[2,6)$, and $[8, 24)$ during evaluations. The distributions of node numbers remain the same. As claimed, our models successfully generalize to graphs of different magnitudes with far better performance than baselines. Notably, the Homo-Path model even achieves the same performance (relative loss $\approx 0.03$) for all scales of magnitudes, which experientially supports Theorem~\ref{thm: homo-generalize}. The Iter-Homo-Path model performs slightly worse because the sigmoid function in the iterative module is not homogeneous.

\textbf{Interpreting stopping criterion learned by the iterative module.}
\label{sec:exp.interpretable}
We show that our \mbox{Iter-Homo-Path} model learned the optimal stopping criterion for the unweighted shortest path problem in Figure \ref{fig:iteration-number}. 
Typically, to accurately predict the shortest path of lengths $d$ on undirected graphs, the iteration number of GNN layers is at least $d/2$ due to the message passing nature of GNNs (see Appendix for details). 
Our iterative module learned such optimal stopping criterion. The Iter-Homo-Path model adaptively increases the iteration numbers w.r.t. the distances and, moreover, stops timely when the information is enough. 

\subsection{General reasoning tasks}
\label{sec:exp.general-reasoning}

\textbf{Physical simulation.}
\label{sec:exp.physical-simulation}
We evaluate the generalizability of our models by predicting the moving patterns between objects in a physical simulator. We consider an environment 
called \textit{Newton's ball}:
all balls with the same mass lie on a friction-free pathway. With the ball at one end moving towards others, our model needs to predict the motion of the balls of both ends at the next time step. 
The metric is the mean squared error. Models are trained in worlds with $[4,34)$ balls and are tested in worlds with 100 balls. 
As shown in Table \ref{tab:summary}, the Iter-Homo-Path model and the Iter-Path model significantly outperform others, demonstrating the advantages of our iterative module for improving generalizability w.r.t. scales. The homogeneous prior is not as beneficial
since the target functions are not homogeneous.

\textbf{Symbolic PacMan.}
\label{sec:exp.pacman}
To show that our iterative module can improve reinforcement learning, we construct a symbolic PacMan environment with similar rules to the PacMan in Atari ~\cite{bellemare13arcade}. 
The environment contains a map with dots and walls. 
The agent needs to figure out a policy to quickly ``eat'' all dots while avoiding walls on the map to maximize the return. We abstract the observations as graphs using the landmark~\cite{huang2019mapping}. 
We adopt Double Q learning~\cite{van2016deep} to train the policy.
Unlike original Atari PacMan, 
our environment is more challenging because we randomly sample the layout of maps for each episode, and we test models in environments with different numbers of dots and walls.
The agent cannot just remember one policy to be successful but needs to
learn to do planning according to the current observation.
The metric is the success rate of eating dots. Our IterGNN (97.5\%) performs much better than baselines, CNN (91.5\%) and PointNet~\cite{qi2017pointnet} (29.0\%). 
Our IterGNN also shows remarkable generalizability among different environment settings.
For example, even though the models are trained in environments with 10 dots and 8 walls, our IterGNN achieves a 94.0\% success rate in environments with 10 dots and 15 walls
and 93.4\% in environments with 8 walls and 20 dots. The tables that list the generalization performance of IterGNN, GCN, and PointNet in 30 different settings of environments are presented in the Appendix to save space.

\textbf{Image-based navigation.}
\label{sec:exp.image-navi}
We show the benefits of the differentiability of a generalizable reasoning module using the image-based navigation task.
The model needs to plan the shortest route from the source to target on 2D images with obstacles. However, the properties of obstacles are not given as a prior, and the model must discover them based on image patterns during training. 
We simplify the task by defining each pixel as obstacles merely according to its own pixel values.
As stated in Table~\ref{tab:summary}, our Iter-Homo-Path model successfully solves the task. The model achieves success rates larger than 90\% for finding the shortest paths on images of size $16\times 16$, and $33\times 33$, while it is only trained on images of size $\le10\times 10$. All of our proposals help improve generalizability.

\section{Conclusion}

We propose an iterative module and the homogeneous prior to improve the generalizability of GNNs w.r.t. graph scales. Experiments show that our proposals do improve the generalizability for solving multiple graph-related problems and tasks.

\section{Acknowledgements}
H. Tang and B. -L. Lu were supported in part by the National Key Research and Development Program of China (2017YFB1002501), the National Natural Science Foundation of China (61673266 and 61976135), SJTU Trans-med Awards Research (WF540162605), the Fundamental Research Funds for the Central Universities, and the 111 Project. H. Su, Z. Huang, and J. Gu were supported by the NSF grant IIS-1764078. We specially thank Zhizuo Zhang, Zhiwei Jia, and Chutong Yang for the useful discussions, and Wei-Long Zheng, Bingyu Shen and Yuming Zhao for reviewing the paper prior to submission.

\section{Broader Impact}

Our methods provide general tools to improve the generalizability of GNNs with respect to scales. This work can thus be applied to many applications of GNNs, such as natural language processing, traffic prediction, and recommendation systems. They have many potential positive impact in the society. For example, better traffic prediction enables shorter traffic time for all vehicles, which could help protect the environment. Improved recommendation system could promote the transition of information for more productivity and more fairness. Moreover, by improving the generalizability with respect to scales, models can be trained on graphs of much smaller scales than reality. It reduces the cost of collecting and storing large datasets with large samples, which can then alleviate the risks of violating privacy and of 
harming the environment. 
On the other hand, this work may also have negative consequences. Improving techniques in the field of natural language processing can help monitor and collect personal information of each individual. Stronger recommendation system can also hurt the fairness as different information targeted to different groups of people.

\bibliography{neurips_2020}
\bibliographystyle{neurips}


	\appendix
	
	\section*{\huge{Appendices}}
	
	\section{Organization of the Appendices}
	
	In the supplementary materials, we aim to answer the following questions: \textbf{(1)} How powerful is our iterative module on approximating the iterative algorithms? \textbf{(2)} Are low generalization errors achievable when using homogeneous neural networks to approximate the homogeneous functions? What is the generation error bound? \textbf{(3)} Are low training errors achievable when using HomoMLP to approximate the homogeneous functions? Is HomoMLP a universal approximator of positive homogeneous functions? \textbf{(4)} Is the iterative module harmful to the standard generalizability in practice? What is its performance on graph-classification benchmarks? \textbf{(5)} What are the experimental setups? How are the models built?
	
	To answer the question \textbf{(1)}, we present the theoretical analysis of the representation power of our iterative module in Section~\ref{sec:theory-iterative}. To answer the question \textbf{(2)}, we prove the generalization error bounds of homogeneous neural networks on approximating homogeneous functions with independent scaling assumptions in Section~\ref{sec:theory-homo-generalize}. A concrete bound based on the PAC-Bayesian framework is also presented in Lemma~\ref{lemma:homo-generalize-bound}. To answer the question \textbf{(3)}, we present and prove the universal approximation theorem on approximating the homogeneous functions for both general HomoMLP and width-bounded HomoMLP in Section~\ref{sec:theory-homo-power}. To answer question \textbf{(4)}, we show that our IterGIN model, which wraps each layer of the state-of-art GIN~\cite{xu2018how} model with our iterative module, achieves competitive performance to GIN, in Section~\ref{sec:exp-graph-classification}. To answer question \textbf{(5)}, we describe all omitted details of the experimental setups in Section~\ref{sec:exp-setups}. 
	
	Moreover, we provide the index of contents following the same order as they appear in the main body of the paper, as follows:
	
	\begin{itemize}
		\item Details of IterGNN, such as the memory-efficient implementation in Section~\ref{sec:iterative-memory-efficient}, the
		theoretical analysis of representation powers in Section~\ref{sec:theory-iterative}, the node-wise iterative module to support unconnected graphs in Section~\ref{sec:iterative-node-wise}, and the decaying confidence mechanism to achieve much larger iteration numbers during inference in practice in Section~\ref{sec:iterative-decay}.
		\item Theoretical analysis of homogeneous neural networks, including the proof of Theorem~\ref{thm: homo-generalize} in Section~\ref{sec:theory-homo-generalize}, the proof of Theorem~\ref{thm:theorem2} in Section~\ref{sec:theory-homo-power}, the propositions stating that HomoGNN and HomoMLP can only represent homogeneous functions in Section~\ref{sec:theory-homo-homo}.
		\item The formulation details of PathGNN layers in Section~\ref{appsec:pathgnn}.
		\item The generation processes and the properties of datasets in Section~\ref{sec:exp-setups}.
		\item The details of models and baselines in Section~\ref{sec:exp.models}.
		\item The training details in Section~\ref{sec:training-details}.
		\item The ACT algorithm usually learns small iteration numbers in Section~\ref{sec:exp-result-iter-num}.
		\item The minimum depth of GNNs to accurately predict the shortest path of length $l$ is $l/2$ in Section~\ref{sec:exp-result-iter-num}.
		\item The generalization performance of IterGNN, GCN, and PointNet on the symbolic Pacman task in environments with different number of dots and different number of walls in Section~\ref{sec:exp-results-pacman}.
	\end{itemize}
	
	In general, we provide the theoretical analysis of our proposals in Section~\ref{sec:theory}. We describe the detailed formulations of our proposals in Section~\ref{sec:method}. The omitted experimental setups are all listed in Section~\ref{sec:exp-setups} and the omitted experimental results are presented in Section~\ref{sec:exp-results}. At last, we also state more background knowledges of graph neural networks (GNNs) in Section~\ref{sec:background}.
	
	\section{Theoretical analysis}
	\label{sec:theory}

	We present the theoretical analysis of our proposals in this section. Main results include
	\begin{itemize}
		\item The representation powers of our iterative module. \begin{itemize}
			\item Our iterative module is a universal approximator of the iterative algorithms, with oracles to reproduce the body function and the condition function in the iterative algorithms. (Theorem~\ref{thm:iter-power})
			\item Under some more practical assumptions, we show that our iterative module can achieve adaptive and unbounded iteration numbers depending on the graph sizes using GNNs. (Proposition~\ref{prop:iter-power})
		\end{itemize}
		\item Generation error bounds of homogeneous neural networks.
		\begin{itemize}
			\item We prove that the generation error bounds of homogeneous neural networks on approximating the homogeneous functions scale linearly with the expectation of the scales/magnitudes under the independent scaling assumption. (Theorem~\ref{thm: homo-generalize})
			\item We provide a concrete generation error bounds for homogeneous neural networks by integrating Theorem~\ref{thm: homo-generalize} and a specific generation error bounds with classical i.i.d. assumptions in the PAC-Bayesian framework (Eq.7 in~\cite{NIPS2017_7176}). (Lemma~\ref{lemma:homo-generalize-bound})
		\end{itemize}
		\item The homogeneous properties of HomoMLP and of HomoGNN.
		\begin{itemize}
			\item We prove that HomoMLP and HomoGNN can only represent homogeneous functions. (Proposition~\ref{prop:homo-prop-homomlp} and Proposition~\ref{prop:homo-prop-homognn})
		\end{itemize}
		\item The universal approximation theorems of homogeneous functions for HomoMLP.
		\begin{itemize}
			\item We prove that HomoMLP is a universal approximator of homogeneous functions. (Theorem~\ref{thm:homo-univ-homomlp})
			\item We prove that width-bounded HomoMLP is also a universal approximator of homogeneous functions. (Theorem~\ref{thm:theorem2} and Theorem~\ref{thm:homo-univ-bounded-homomlp})
		\end{itemize}
	\end{itemize}
	
	\subsection{Representation powers of iterative module}
	\label{sec:theory-iterative}
	
	We first state the intuition that our iterative module as described in the main body can approximate any iterative algorithms as defined in Algorithm \ref{alg:iteration_algorithm}, as long as the $body$ and $condition$ functions are available or can be perfectly reproduced by neural networks.
	
	\begin{algorithm}
		\caption{Iterative algorithm}
		\label{alg:iteration_algorithm}
		\begin{algorithmic}
			\STATE  {\bfseries input } initial feature $x$
			\STATE $k\gets 1$
			\STATE $h^0\gets x$
			\WHILE{not $condition(h^k)$}
			\STATE $h^k\gets body(h^{k-1})$
			\STATE $k\gets k+1$
			\ENDWHILE
			\STATE  {\bfseries return }$h=h^k$
		\end{algorithmic}
	\end{algorithm}
	
	More formally, we build an ideal class of models, named as Iter-Oracle, by combining our iterative module with the oracles $\mathcal{F}_\theta$ that can perfectly reproduce the $body$ function and the $condition$ function, which means there exist $\theta'$ and $\theta''$ such that for all $x$,
	$\mathcal{F_{\theta'}}(x) = body(x)$ and 
	$\mathcal{F_{\theta''}}(x) = condition(x)$. We can then show the representation power of our iterative module using the following theorem:
	\begin{appmytheorem}
		For any iterative algorithm, $\text{iter-alg}$, defined as in Algorithm~\ref{alg:iteration_algorithm}, for any initial feature $x$, and for any $\epsilon>0$, there exist an Iter-Oracle model $\mathcal{A}$, which represents the function $\mathcal{F}_\mathcal{A}$, satisfying 
		\begin{eqnarray}
		||\text{iter-alg}(x)-\mathcal{F}_\mathcal{A}(x)|| < \epsilon.
		\label{eq:theorem-iter-power}
		\end{eqnarray}
		\label{thm:iter-power}
	\end{appmytheorem}
	We prove it by construction. We build the function $f$ in our iterative module using $\mathcal{F}_{\theta'}$ such that $\mathcal{F}_{\theta'}(x) = body(x)$ for all $x$. The function $g$ in our iterative module is built as $\text{sigmoid}(\alpha(\mathcal{F}_{\theta''}(x)-0.5))$, where $\mathcal{F}_{\theta''}(x)=condition(x)$ for all $x$ and we utilize similar rules to the python language for type conversion, which means $condition(x)$ outputs $1$ if the condition is satisfied and outputs $0$ otherwise. By setting $\alpha\rightarrow +\infty$, we have 
	\begin{equation*}
	c^k\rightarrow\begin{cases}
	1 & \text{if}~condition(h^k) \\
	0 & \text{if not}~condition(h^k).
	\end{cases}
	\end{equation*}
	Therefore, $\sum_{j=1}^{\infty}c^jh^j\prod_{i=1}^{j-1}(1-c^i)\rightarrow h^k$, where $k$ is the time step that $condition(h^k)$ is firstly satisfied. More formally, assume $\Lambda$ bounds the norm of features $h^j$ for the specific iterative algorithm $iter\text{-}alg$, for the specific initial feature $x$, and for any $j$, we can set $\alpha$ as 
	\begin{eqnarray}
	\alpha > 2\ln\left(\frac{(k+1)\Lambda}{\epsilon}-1\right) \text{ and } \alpha > -2\ln\left(\left(1+\frac{\epsilon}{(k+1)\Lambda}\right)^{-\frac{1}{k}}-1\right),
	\end{eqnarray}
	so that Eq.~\ref{eq:theorem-iter-power} is satisfied, since for $j<k$, 
	\begin{eqnarray}
	\left\lVert c^jh^j\prod_{i=1}^{j-1}(1-c^i)-0\right\rVert < \left\lVert c^jh^j\right\rVert < \frac{1}{1+e^{\frac{\alpha}{2}}}\Lambda = \frac{\epsilon}{k+1},
	\end{eqnarray}
	for $j=k$,
	\begin{eqnarray}
	\left\lVert c^kh^k\prod_{i=1}^{k-1}(1-c^i)-h^k\right\rVert < \left(\left(\frac{1}{1+e^{-\frac{\alpha}{2}}}\right)^{k}-1\right)\Lambda = \frac{\epsilon}{k+1},
	\end{eqnarray}
	for $j>k$,
	\begin{eqnarray}
	\left\lVert\sum_{j=k+1}^\infty c^jh^j\prod_{i=1}^{j-1}(1-c^i)-0\right\rVert < (1-c^k)\Lambda < \frac{1}{1+e^{\frac{\alpha}{2}}}\Lambda = \frac{\epsilon}{k+1}.
	\end{eqnarray}
	Together, we have 
	\begin{eqnarray}
	\left\lVert\sum_{j=1}^\infty c^jh^j\prod_{i=1}^{j-1}(1-c^i)-h^k\right\rVert & \le & \sum_{j=1}^{k-1}\left\lVert c^jh^j\prod_{i=1}^{j-1}(1-c^i)-0\right\rVert + \\
	& & \left\lVert c^kh^k\prod_{i=1}^{k-1}(1-c^i)-h^k\right\rVert + \\
	& &  \left\lVert\sum_{j=k+1}^\infty c^jh^j\prod_{i=1}^{j-1}(1-c^i)-0\right\rVert \\
	& < & (k+1)\frac{\epsilon}{k+1} = \epsilon
	\qed \nonumber
	\end{eqnarray}
	
	We then derive a more practical proposition of IterGNN, based on Theorem~\ref{thm:iter-power}, stating the intuition that IterGNN can achieve adaptive and unbounded iteration numbers:
	\begin{proposition}
		Under the assumptions that IterGNN can calculate graph sizes $N$ with no error and the multilayer perceptron used by IterGNN is a universal approximator of continuous functions on compact subsets of $\mathbb{R}^n$ ($n\ge1$) (i.e. the universal approximation theorem), there exist IterGNNs whose iteration numbers are constant, linear, polynomial or exponential functions of the graph sizes.
		\label{prop:iter-power}
	\end{proposition}
	
	The proofs are simple. Let $g'$ be the function that maps the graph sizes $N$ to iteration numbers $k$. we can then build GNNs as $f$ to calculate the graph sizes $N$ and the number of current time step $j$, and build $g$ as $\text{sigmoid}(\alpha(0.5-|g'(N)-j|))$. The $\alpha$ can be set similarly to the previous proof except for one scalar that compensates the approximation errors of neural networks. More formally, assume $\epsilon'$ bounds the error of predicting $g'(N)$ and $j$ using neural networks $\mathcal{A}$, which means 
	for any input, there exists neural networks that represent functions $\mathcal{F}_{g}$ and $\mathcal{F}_j$ satisfying $|g'(N)-\mathcal{F}_g|<\epsilon'$ and $|j-\mathcal{F}_j|<\epsilon'$. We can then set $\alpha$ as 
	\begin{eqnarray}
	\alpha > \frac{2}{1-4\epsilon'}\ln\left(\frac{(k+1)\Lambda}{\epsilon}-1\right) \text{ and } \alpha > -\frac{2}{1-4\epsilon'}\ln\left(\left(1+\frac{\epsilon}{(k+1)\Lambda}\right)^{-\frac{1}{k}}-1\right),
	\end{eqnarray}
	so that Proposition~\ref{prop:iter-power} is satisfied. Note that it is easy to build GNNs to exactly calculate the graph sizes $N$ and the number of current time step $j$. Given the universal approximation theorem of MLP, the stopping condition function $g$ can also be easily approximated by MLPs. Our iterative module is thus able to achieve adaptive and unbounded iteration numbers using neural networks.

	\subsection{Homogeneous prior}
	\label{sec:theory-homogeneous}
	
	We formalize the generalization error bounds of homogeneous neural networks on approximating homogeneous functions under proper conditions, by extending the example in the main body to more general cases, in Section~\ref{sec:theory-homo-generalize}. 
	To make sure functions represented by neural networks are homogeneous, 
	we also prove that HomoGNN and HomoMLP can only represent homogeneous functions, in Section~\ref{sec:theory-homo-homo}.
	To show that low training errors are achievable, we further 
	analyze the representation powers of HomoMLP and 
	demonstrate that it is a universal approximator of homogeneous functions under some assumptions, based on the universal approximation theorem for width-bounded ReLU networks~\cite{NIPS2017_7203}, in Section~\ref{sec:theory-homo-power}.
	
	\subsubsection{Proof of Theorem 1: generalization error bounds of homogeneous neural networks}
	\label{sec:theory-homo-generalize}
	
	Extending the example in the main body to more general cases, we present the out-of-distribution generalization error bounds of homogeneous neural networks on approximating homogeneous functions under the assumption of independent scaling of magnitudes during inference:
	
	Let training samples $D_m=\{x_1, x_2, \cdots x_m\}$ be independently sampled from the distribution $\mathcal{D}_x$, then if we scale the training samples with the scaling factor $\lambda \in \mathbb{R}^+$ which is independently sampled from the distribution $\mathcal{D}_\lambda$, we get a ``scaled'' distribution $\mathcal{D}_x^\lambda$, which has a  density function $P_{\mathcal{D}_x^\lambda}(z) := \int_{\lambda}\int_{x} \delta(\lambda x=z)P_{\mathcal{D}_\lambda}(\lambda)P_{\mathcal{D}_x}(x)\dif x \dif\lambda$. The following theorem bounds the generalization error bounds on $\mathcal{D}_x^{\lambda}$: 
	\begin{apptheorem} \upshape{(Generalization error bounds of homogeneous neural networks with independent scaling assumption)}. \itshape 
		For any positive homogeneous functions function $f$ and neural network $F_\mathcal{A}$, let $\beta$ bounds the generalization errors on the training distribution $D_x$ , i.e.,  
		$\mathbb{E}_{x\sim\mathcal{D}_x}|f(x)-\mathcal{F}_\mathcal{A}(x)| \le \frac{1}{m}\sum_{i=1}^m |f(x_i)-F_\mathcal{A}(x_i)|+\beta$, then the generalization errors on the scaled distributions $\mathcal{D}_x^\lambda$ scale linearly
		with the expectation of scales $\mathbb{E}_{\mathcal{D}_\lambda}[\lambda]$:
		\begin{eqnarray}
		\mathbb{E}_{x\sim\mathcal{D}_x^\lambda}|f(x)-F_\mathcal{A}(x)|= \mathbb{E}_{\mathcal{D}_\lambda}[\lambda] \mathbb{E}_{x\sim\mathcal{D}_x} |f(x)-F_\mathcal{A}(x)| \le \mathbb{E}_{\mathcal{D}_\lambda}[\lambda](\frac{1}{m}\sum_{i=1}^m |f(x_i)-F_\mathcal{A}(x_i)|+\beta)
		\end{eqnarray}
		\label{appthm: homo-generalize}
	\end{apptheorem}
	The proof is as simple as re-expressing the formulas:
	\begin{eqnarray}
	\mathbb{E}_{z\sim\mathcal{D}_x^\lambda}|f(z)-F_\mathcal{A}(z)| & = & \int_{\lambda}\int_{x} P_{\mathcal{D}_\lambda}(\lambda)P_{\mathcal{D}_x}(x)|f(\lambda x)-F_\mathcal{A}(\lambda x)|\dif x \dif\lambda \\
	& = & \int_{\lambda}P_{\mathcal{D}_\lambda}(\lambda)\lambda \dif\lambda \int_{x}P_{\mathcal{D}_x}(x)|f( x)-F_\mathcal{A}(x)|\dif x \\
	& = & \mathbb{E}_{\mathcal{D}_\lambda}[\lambda] \mathbb{E}_{x\sim\mathcal{D}_x} |f(x)-F_\mathcal{A}(x)| \\
	& \le & \mathbb{E}_{\mathcal{D}_\lambda}[\lambda](\frac{1}{m}\sum_{i=1}^m |f(x_i)-F_\mathcal{A}(x_i)|+\beta)
	\nonumber\qed
	\end{eqnarray}
	
	Theorem~\ref{thm: homo-generalize} can be considered as a meta-bound and can thus be integrated with any specific generalization error bounds with classical i.i.d. assumptions to create a concrete generation error bounds of homogeneous neural networks on approximating homogeneous functions with independent scaling assumptions. For example, when integrated a generation error bounds in the PAC-Bayesian framework (Eq.7 in~\cite{NIPS2017_7176}), we obtain the following lemma: Let $f_\textit{\textbf{w}}$ be any predictor learned from training data. We consider a distribution $\mathcal{Q}$ over predictors with weights of the form $\textit{\textbf{w}}+\textit{\textbf{v}}$, where $\textit{\textbf{w}}$ is a single predictor learned from the training set, and $\textit{\textbf{v}}$ is a random variable. 
	
	\begin{applemma}
		Assume all hypothesis $h$ and $f_{\textbf{w}+\textbf{v}}$ for any $\textbf{v}$ are positive homogeneous functions, as defined in Definition 1. Then, given a “prior” distribution $P$ over the hypothesis that is independent of the training data, with probability at least $1-\delta$ over the draw of the training data, the expected error of $f_{\textbf{w}+\textbf{v}}$ on the scaled distribution $\mathcal{D}_x^\lambda$ can be bounded as follows
		\begin{eqnarray}
		\mathbb{E}_\textbf{v}\left[\mathbb{E}_{x\sim\mathcal{D}_x^\lambda}\left[\left\lvert f(x)-f_{\textbf{w}+\textbf{v}}(x)\right\rvert\right]\right] & \le & \mathbb{E}_{\mathcal{D}_\lambda}\left[\lambda\right]
		\left(
		\mathbb{E}_{\textbf{v}}\left[\frac{1}{m}\sum_{i=1}^m \left\lvert f(x_i)-f_{\textbf{w}+\textbf{v}}(x_i)\right\rvert\right]\nonumber
		+
		4\sqrt{\frac{1}{m}\left(KL(\textbf{w}+\textbf{v}||P)+\ln\frac{2m}{\delta}\right)}\vphantom{\frac12}\right).
		\end{eqnarray}
		\label{lemma:homo-generalize-bound}
	\end{applemma}
	
	\subsubsection{HomoMLP and HomoGNN are homogeneous functions}
	\label{sec:theory-homo-homo}
	
	We present that HomoGNN and HomoMLP can only represent homogeneous functions:
	
	\begin{appproposition}
		For any input $x$, we have $\text{HomoMLP}(\lambda x) = \lambda \text{HomoMLP}(x)$ for all $\lambda>0$.
		\label{prop:homo-prop-homomlp}
	\end{appproposition}
	
	\begin{appproposition}
		For any graph $G=(V,E)$ with node attributes $\vec{x}_v$ and edge attributes $\vec{x}_e$, we have for all $\lambda>0$,
		\begin{eqnarray}
		\text{HomoGNN}(G, \{\lambda\vec{x}_v:v\in V\}, \{\lambda\vec{x}_e:e\in E\}) = \lambda \text{HomoGNN}(G, \{\vec{x}_v:v\in V\}, \{\vec{x}_e:e\in E\}).
		\end{eqnarray}
		\label{prop:homo-prop-homognn}
	\end{appproposition}
	
	Both propositions are derived from the following lemma:
	
	\begin{applemma}
		Compositions of homogeneous functions are homogeneous functions. 
		\label{lemma:homo-prop}
	\end{applemma}
	We compose functions by taking the outputs of functions as the input of other functions. The inputs of functions are either the outputs of other functions or the initial features $x$. For example, we can compose functions $f,g,h$ as $h(f(x),g(x), x)$. If $f,g,h$ are all homogeneous functions, we have for all $x$ and all $\lambda>0$,
	\begin{eqnarray}
	h(f(\lambda x),g(\lambda x), \lambda x) = \lambda h(f(x),g(x), x).
	\end{eqnarray}
	More formally, we can prove the lemma by induction. We denote the composition of a set of functions $\{f_1, f_2, \cdots, f_n\}$ as $O(\{f_1, f_2, \cdots, f_n\})$. Note that there are multiple ways to compose $n$ functions. Here, $O(\{f_1, f_2, \cdots, f_n\})$ just denotes one specific way of composition, and we can use $O'(\{f_1, f_2, \cdots, f_n\})$ to denote another. We want to prove that the composition of homogeneous functions is still a homogeneous function, which means for all $n>0$, for all $\lambda > 0$, and for all possible ways of compositions $O$, $O(\{f_1, f_2, \cdots, f_n\})(\lambda x) = \lambda O(\{f_1, f_2, \cdots, f_n\})(x)$.

	The base case: $n=1$. The composition of a single function $O(\{f_1\})$ is itself $f_1$. Therefore,  $O(\{f_1\})$ is homogeneous by definition as $O(\{f_1\})(\lambda x) = f_1(\lambda x) = \lambda O(\{f_1\})(x)$.
	
	Assume the composition of the $k$ functions is homogeneous, for any composition of $k+1$ functions $O(\{f_1, f_2, \cdots, f_{k+1}\})$, let $f_{k+1}$ be the last function of the compositions, which means $O(\{f_1, f_2, \cdots, f_{k+1}\})(x) := f_{k+1}(O'(\{f_1, f_2, \cdots, f_k\})(x), O''(\{f_1, f_2, \cdots, f_k\})(x), O'''(\{f_1, f_2, \cdots, f_k\})(x), \cdots)$, it's easy to see that, according to the definition of homogeneous functions, 
	\begin{eqnarray}
	O(\{f_1, f_2, \cdots, f_{k+1}\})(\lambda x) = \lambda O(\{f_1, f_2, \cdots, f_{k+1}\})(x).
	\qed\nonumber
	\end{eqnarray} 
	
	Therefore, we prove the Lemma~\ref{lemma:homo-prop}. Proposition~\ref{prop:homo-prop-homomlp} and Proposition~\ref{prop:homo-prop-homomlp} can all be considered as specializations of this lemma.
	
	\subsubsection{Proof of Theorem 2: representation powers of HomoMLP}
	\label{sec:theory-homo-power}
	
	We first introduce a class of neural networks (defined in Eq.\ref{eq:homo-norm}) that can be proved as a universal approximator of homogeneous functions (Theorem~\ref{thm:homo-univ-gmlp}) and then show that our HomoMLP is as powerful as this class of neural networks to prove that our HomoMLP is a universal approximator of homogeneous functions (Theorem~\ref{thm:homo-univ-homomlp}). Furthermore, we prove the universal approximation theorem for width-bounded HomoMLP (Theorem~\ref{thm:theorem2}). At last, we notice that the assumption in Theorem~\ref{thm:theorem2} (i.e. ``for all homogeneous and Lebesgue-integrable functions: $f:\mathbb{R}^n\rightarrow\mathbb{R}$'') is too strong. Most common homogeneous functions are not Lebesgue-integrable functions on $\mathbb{R}^n$ (e.g. $f(x)=||x||$ is a not Lebesgue-integrable function as its integration on $\mathbb{R}^n$ is infinite.). We therefore prove another universal approximation theorem for width-bounded HomoMLP with assumptions that are more reasonable both theoretically and practically (Theorem~\ref{thm:homo-univ-bounded-homomlp}).
	
	We construct a class of neural networks as the universal approximator of positive homogeneous functions, as defined in the main body, as follows:
	\begin{equation}
	G_{\text{MLP}}(x) = |x|\mathcal{F}_\text{MLP}(\frac{x}{|x|}),
	\label{eq:homo-norm}
	\end{equation}
	where $|\cdot|$ denotes the L1 norm, the MLP denotes the classical multilayer perceptrons with positive homogeneous activation functions and $\mathcal{F}_\text{MLP}$ is the function represented by the specific MLP. 
	
	\begin{appproposition}
		For any input $x$, we have $G_{\text{MLP}}(\lambda x) = \lambda G_{\text{MLP}}(x)$.
	\end{appproposition}
	
	This proposition can be easily proved by re-expressing the formulas: 
	\begin{eqnarray}
	G_{\text{MLP}}(\lambda x) = |\lambda x|\mathcal{F}_\text{MLP}(\frac{\lambda x}{|\lambda x|}) = \lambda |x|\mathcal{F}_\text{MLP}(\frac{x}{|x|}) = \lambda G_{\text{MLP}}(x)
	\nonumber\qed
	\end{eqnarray}
	
	We show that it is a universal approximator of homogeneous functions: Let $\mathbb{X}$ be a compact subset of $\mathbb{R}^m$  and $C(\mathbb{X})$ denotes the space of real-valued continuous functions on $\mathbb{X}$.
	
	\begin{appmytheorem}
		\upshape(Univeral approximation theorem for $G_{\text{MLP}}$.)~\itshape
		Given any $\epsilon>0$ and any function $f\in C(\mathbb{X})$, there exist a finite-layer feed-forward neural networks $\mathcal A$ with positive homogeneous activation functions such that for all $x\in \mathbb{X}$, we have $|G_\mathcal{A}(x)-f(x)|=\left\lvert\left\lvert x\right\rvert \mathcal{F}_\mathcal{A}\left(\frac{x}{\left\lvert x\right\rvert}\right)-f\left(x\right)\right\rvert<\epsilon$.
		
		\label{thm:homo-univ-gmlp}
	\end{appmytheorem}

	The prove is as simple as applying the universal approximation theorem of MLPs~\cite{cybenko1989approximation} and applying the definition of the homogeneous functions. In detail, as $\mathbb{X}$ is a compact subset, the magnitudes of inputs $x$ is bounded. We use $M$ denote the bound, which means $\left\lvert x\right\rvert \le M$ for all $x\in \mathbb{X}$. According to the universal approximation theorem of MLPs~\cite{cybenko1989approximation}, there exists a finite-layer feed-forward layer $\mathcal{A}$ with ReLU as activation functions such that $\left\lvert \mathcal{F}_\mathcal{A}\left(\frac{x}{\left\lvert x\right\rvert}\right)-f\left(\frac{x}{\left\lvert x\right\rvert}\right)\right\rvert<\frac\epsilon M$ for all $x\in \mathbb{X}$. Then, according to definition of homogeneous functions, we have 
	\begin{eqnarray}
	|G_\mathcal{A}(x)-f(x)|=\left\lvert\left\lvert x\right\rvert \mathcal{F}_\mathcal{A}\left(\frac{x}{\left\lvert x\right\rvert}\right)-\left\lvert x\right\rvert f\left(\frac{x}{\left\lvert x\right\rvert}\right)\right\rvert<\left\lvert x\right\rvert\frac\epsilon M <\epsilon
	\end{eqnarray}
	Note that ReLU is positive homogeneous. Therefore, we finish the proof of Theorem~\ref{thm:homo-univ-gmlp}.
	\qed

	We then prove that HomoMLP is a universal approximator of homogeneous functions: Let $\mathbb{X}$ be a compact subset of $\mathbb{R}^m$  and $C(\mathbb{X})$ denotes the space of real-valued continuous functions on $\mathbb{X}$.
	
	\begin{appmytheorem}
		\upshape(Universal approximation theorem for HomoMLP.)~\itshape
		Given any $\epsilon>0$ and any function $f\in C(\mathbb{X})$, there exist a finite-layer HomoMLP $\mathcal A'$, which represents the function $F_{\mathcal{A}'}$, such that for all $x\in \mathbb{X}$, we have $|F_{\mathcal{A}'}(x)-f(x)|<\epsilon$.
		\label{thm:homo-univ-homomlp}
	\end{appmytheorem}
	
	We prove it based on Theorem~\ref{thm:homo-univ-gmlp} by construction. In detail, according to Theorem~\ref{thm:homo-univ-gmlp}, there exists a finite-layer MLP $\mathcal{A}$ such that for all $x\in \mathbb{X}$, $|G_\mathcal{A}(x)-f(x)|=\left\lvert\left\lvert x\right\rvert \mathcal{F}_\mathcal{A}\left(\frac{x}{\left\lvert x\right\rvert}\right)-f\left(x\right)\right\rvert<\epsilon$.
	Without loss of generality, we assume $\mathcal{A}$ as a two-layer ReLU feed-forward neural networks, which means $\mathcal{F}_\mathcal{A}(x) = W^2\text{ReLU}(W^1x+b^1)+b^2$, where $W^1\in \mathbb{R}^{n\times m},W^2\in\mathbb{R}^{1\times n}$ are weight matrices and $b^1\in \mathbb{R}^n,b^2\in\mathbb{R}$ are biases. The function $G_\mathcal{A}$ can then be expressed as 
	\begin{eqnarray}
	G_\mathcal{A}(x) = |x|\left(W^2\text{ReLU}\left(W^1\frac{x}{|x|}+b^1\right)+b^2\right) = W^2\text{ReLU}\left(W^1x+b^1|x|\right)+b^2|x|.
	\label{eq:proof.theorem.c.2.2}
	\end{eqnarray}
	Therefore, we just need to prove that the L1 norm $|\cdot|$ can be exactly calculated by HomoMLP to show that HomoMLP is a universal approximator homogeneous functions. Typically, we construct a two-layer HomoMLP as follows
	\begin{eqnarray}
	\textbf{1}^T\text{ReLU}\left(\begin{bmatrix}
	\textbf{I}\\
	-\textbf{I}
	\end{bmatrix}x\right)\equiv |x|,\text{for all }x\in \mathbb{R}^m,
	\label{eq:proof.theorem.l1norm}
	\end{eqnarray}
	where $\textbf{1}$ is a vertical vector containing $2m$ elements whose values are all one, and $\textbf{I}$ denotes the identity matrix of size $m\times m$. Together with Eq.~\ref{eq:proof.theorem.c.2.2}, we show that $ G_\mathcal{A}(x)$ is a specification of HomoMLP, denoted as $\mathcal{A}'$, as follows:
	\begin{eqnarray}
	G_\mathcal{A}(x) & = & W^2\text{ReLU}\left(W^1x+b^1|x|\right)+b^2|x| \\
	& = & W^2\text{ReLU}\left(W^1x+b^1\textbf{1}^T\text{ReLU}\left(\begin{bmatrix}
	\textbf{I}\\
	-\textbf{I}
	\end{bmatrix}x\right)\right)+b^2\textbf{1}^T\text{ReLU}\left(\begin{bmatrix}
	\textbf{I}\\
	-\textbf{I}
	\end{bmatrix}x\right)\\
	& = & 
	\begin{bmatrix}
	W^2 &
	b^2
	\end{bmatrix}\text{ReLU}\left(
	\begin{bmatrix}
	W^1+b^1 & -W^1+b^1\\
	\textbf{1}^T & \textbf{1}^T
	\end{bmatrix}\text{ReLU}\left(
	\begin{bmatrix}
	\textbf{I}\\ -\textbf{I}
	\end{bmatrix}x
	\right)
	\right)\\
	& = & F_{\mathcal{A}'}(x)
	\end{eqnarray}
	Here, we use $\textbf{1}$ to denote a vertical vector containing $m$ elements whose values are all one, and $\textbf{I}$ still denotes the identity matrix of size $m\times m$. The summation of the weight matrix $W\in \mathbb{R}^{n\times m}$ and the bias $b\in \mathbb{R}^n$ outputs a new weight matrix $W_b$ such that $W_b[i,j]=W[i,j]+b[i]$ for all $i=1,2,\cdots,n$ and $j=1,2,\cdots,m$, where $W[i,j]$ is the element in the $i$th row and the $j$th column of matrix $W$ and $b[i]$ is the $i$th element of vertex $b$. Consequently, we build a three-layer HomoMLP $\mathcal{A}'$ with ReLU as activation functions such that for all $x\in \mathbb{X}$,
	\begin{eqnarray}
	|F_{\mathcal{A}'}(x)-f(x)| = |G_\mathcal{A}(x)-f(x)| < \epsilon.
	\qed\nonumber
	\end{eqnarray}
	
	Applying similar techniques to previous proofs, we can further derive the universal approximation theorem for width-bounded MLP. We first prove Theorem~\ref{thm:theorem2}, which is presented in the main body, and show that it contains a too strong assumption. To amend this, we introduce two more theorems also stating that width-bounded HomoMLP is a universal approximator of positive homogeneous functions under proper conditions.
	
	\begin{apptheorem} \upshape {(Universal approximation theorem for width-bounded HomoMLP).} \itshape
		For any positive-homogeneous Lebesgue-integrable function $f:\mathbb{R}^n\mapsto \mathbb{R}$ and any $\epsilon>0$, there exists a finite-layer HomoMLP $\mathcal{A}$ with width $d_m\le 2(n+4)$, such that the function $F_\mathcal{A}$ represented by this networks satisfies 
		$\int_{\mathbb{R}^n}|f(x)-F_\mathcal{A}(x)|\dif x < \epsilon$.
		\label{thm:theorem2}
	\end{apptheorem}
	
	We notice that the assumption in Theorem~\ref{thm:theorem2} is too strong, because for any positive-homogeneous Lebesgue-integral function $f:\mathbb{R}^n\mapsto \mathbb{R}$, it must satisfy that $\int_{\mathbb{R}^n}|f(x)|\dif x=0$. We prove it by contradiction. Assume there exists a positive-homogeneous Lebesgue-integrable function $f:\mathbb{R}^n\mapsto \mathbb{R}$ such that $\int_{\mathbb{R}^n}|f(x)|\dif x>0$, which means there must exist a bounded subset $\mathbb{E}\in \mathbb{R}^n$ such that $\int_{\mathbb{E}}|f(x)|\dif x>0$. Then, according to the definition of homogeneous functions, we can calculate the integration on the scaled subset $\mathbb{E}_\lambda:=\{\lambda x:x\in E\}$ as $\int_{\mathbb{E}_\lambda}|f(x)|\dif x=\lambda\int_{\mathbb{E}}|f(x)|\dif x$. By setting $\lambda= \infty$, we have $\int_{\mathbb{R}^n}|f(x)|\dif x \ge \int_{\mathbb{E}_\lambda}|f(x)|=\infty$, which contradicts the assumption that $f$ is a Lebesgue-integrable function.
	
	According to this observation, the proof of Theorem~\ref{thm:theorem2} is as simple as setting the weight matrices of HomoMLP as zero, so that $\int_{\mathbb{R}^n}|f(x)-F_\mathcal{A}(x)|\dif x = \int_{\mathbb{R}^n}|f(x)|\dif x = 0 < \epsilon$.\qed
	
	We then change the assumption from ``for any positive-homogeneous Lebesgue-integral function $f:\mathbb{R}^n\mapsto \mathbb{R}$'' to ``for any positive-homogeneous Lebesgue-integral function $f:\mathbb{X}\mapsto \mathbb{R}$, where $\mathbb{X}$ is a Lebesgue-measurable and compact subset of $\mathbb{R}^n$'', so that most popular homogeneous functions  are taken into account in our Theorems. Note that this assumption is also reasonable in practice since numbers stored in our computers are also bounded within a measurable compact big cube $[-M,M]^n$, where $M$ is the biggest number that can be stored in computers. Theorem~\ref{thm:homo-univ-bounded-gmlp} and Theorem~\ref{thm:homo-univ-bounded-homomlp} are proved to show that width-bounded $G_\text{MLP}$ and width-bounded HomoMLP are universal approximators of homogeneous functions with the amended assumption.
	
	\begin{appmytheorem}
		\upshape~{(Universal approximation theorem for width-bounded $G_\text{MLP}$).}~\itshape
		For any positive-homogeneous Lebesgue-integrable function $f:\mathbb{X}\mapsto \mathbb{R}$, where $\mathbb{X}$ is a Lebesgue-measurable compact subset of $\mathbb{R}^n$, and for any $\epsilon>0$, there exists a finite-layer feed-forward neural networks $\mathcal{A}$ with positive homogeneous activation functions and with width $d_m\le n+4$, such that 
		\begin{eqnarray}
		\int_{\mathbb{X}}|f(x)-G_\mathcal{A}(x)|\dif x:=\int_{\mathbb{X}}\left\lvert\left\lvert x\right\rvert \mathcal{F}_\mathcal{A}\left(\frac{x}{\left\lvert x\right\rvert}\right)-f(x)\right\rvert\dif x < \epsilon.
		\end{eqnarray}
		\label{thm:homo-univ-bounded-gmlp}
	\end{appmytheorem}
	
	The proof is very similar to the proof of Theorem~\ref{thm:homo-univ-gmlp}. In detail, as $\mathbb{X}$ is a compact subset, the magnitudes of inputs $x$ is bounded. We use $M$ denote the bound, which means $\left\lvert x\right\rvert \le M$ for all $x\in \mathbb{X}$. According to the universal approximation theorem of width-bounded MLPs~\cite{NIPS2017_7203}, there exists a finite-layer feed-forward layer $\mathcal{A}$ with ReLU as activation functions and with width $d_m<n+4$, such that
	\begin{eqnarray}
	\int_{\mathbb{X}}\left\lvert \mathcal{F}_\mathcal{A}\left(\frac{x}{\left\lvert x\right\rvert}\right)-f\left(\frac{x}{\left\lvert x\right\rvert}\right)\right\rvert\dif x<\frac{\epsilon}{M}.
	\end{eqnarray}
	Then, according to the definition of homogeneous functions, we have 
	\begin{eqnarray}
	\int_{\mathbb{X}}|f(x)-G_\mathcal{A}(x)|\dif x=\int_{\mathbb{X}}\left\lvert\left\lvert x\right\rvert \mathcal{F}_\mathcal{A}\left(\frac{x}{\left\lvert x\right\rvert}\right)-\left\lvert x\right\rvert f\left(\frac{x}{\left\lvert x\right\rvert}\right)\right\rvert\dif x < M\frac{\epsilon}{M} = \epsilon.
	\end{eqnarray}
	Note that ReLU is positive homogeneous. Therefore, we finish the proof of Theorem~\ref{thm:homo-univ-bounded-gmlp}.
	\qed
	
	\begin{appmytheorem}
		\upshape~{(Universal approximation theorem for width-bounded HomoMLP with reasonable assumption).}~\itshape
		For any positive-homogeneous Lebesgue-integrable function $f:\mathbb{X}\mapsto \mathbb{R}$, where $\mathbb{X}$ is a Lebesgue-measurable compact subset of $\mathbb{R}^n$, and for any $\epsilon>0$, there exists a finite-layer HomoMLP $\mathcal{A}'$ with width $d_m\le 2(n+4)$, which represents the function $F_{\mathcal{A}'}$ such that 
		$\int_{\mathbb{X}}|f(x)-F_{\mathcal{A}'}(x)|\dif x < \epsilon$.
		\label{thm:homo-univ-bounded-homomlp}
	\end{appmytheorem}
	
	The proof is very similar to the proof of Theorem~\ref{thm:homo-univ-homomlp}. In detail, according to Theorem~\ref{thm:homo-univ-bounded-gmlp}, there exists a finite-layer MLP $\mathcal{A}$ such that, \begin{eqnarray}
	\int_{\mathbb{X}}|f(x)-G_\mathcal{A}(x)|\dif x=\int_{\mathbb{X}}\left\lvert\left\lvert x\right\rvert \mathcal{F}_\mathcal{A}\left(\frac{x}{\left\lvert x\right\rvert}\right)-f(x)\right\rvert\dif x < \epsilon.
	\end{eqnarray}
	Assume the MLP $\mathcal{A}$ is formulated as 
	\begin{eqnarray}
	\mathcal{F}_\mathcal{A}(x) = W^K\text{ReLU}\left(W^{K-1}\text{ReLU}\left(\cdots \text{ReLU}\left(W^1x+b^1\right)\cdots\right)+b^{K-1}\right)+b^K,
	\end{eqnarray}
	where $K$ is the layer number, $W^1,W^2,\cdots,W^K$ are weight matrices, and $b^1, b^2, \cdots, b^K$ are biases. We can then re-express $G_\mathcal{A}(x)$, using the definition in Eq.~\ref{eq:homo-norm}, as follows
	\begin{eqnarray}
	G_\mathcal{A}(x) = W^K\text{ReLU}\left(W^{K-1}\text{ReLU}\left(\cdots \text{ReLU}\left(W^1x+b^1|x|\right)+\cdots\right)+b^{K-1}|x|\right)+b^K|x|.
	\end{eqnarray}
	Together with Eq.~\ref{eq:proof.theorem.l1norm}, we show that $ G_\mathcal{A}(x)$ is a specification of HomoMLP, denoted as $\mathcal{A}'$, as follows:
	\begin{eqnarray*}
		G_\mathcal{A}(x) & = & W^K\text{ReLU}\left(W^{K-1}\text{ReLU}\left(\cdots \text{ReLU}\left(W^1x+b^1|x|\right)+\cdots\right)+b^{K-1}|x|\right)+b^K|x| \\
		& = & 
		\begin{bmatrix}
			W^K & b^K
		\end{bmatrix}\text{ReLU}(
		\begin{bmatrix}
			W^{K-1} & b^{K-1}\\
			\textbf{0}^T & \textbf{1}^T
		\end{bmatrix}\text{ReLU}( \cdots
		\begin{bmatrix}
			W^2 & b^2\\
			\textbf{0}^T & \textbf{1}^T
		\end{bmatrix}\text{ReLU}( \\
		& &
		\begin{bmatrix}
			W^1+b^1 & -W^1+b^1\\
			\textbf{1}^T & \textbf{1}^T
		\end{bmatrix}\text{ReLU}(
		\begin{bmatrix}
			\textbf{I}\\ -\textbf{I}
		\end{bmatrix}x
		)
		)\cdots
		)
		)
		\\
		& = & F_{\mathcal{A}'}(x).
	\end{eqnarray*}
	Here, we also use $\textbf{1}$ to denote vectors full of ones, $\textbf{0}$ to denote vectors full of zeros, and $\textbf{I}$ to denote the identity matrix of size $m\times m$. The summation of the weight matrix $W\in \mathbb{R}^{n\times m}$ and the bias $b\in \mathbb{R}^n$ outputs a new weight matrix $W_b$ such that $W_b[i,j]=W[i,j]+b[i]$ for all $i=1,2,\cdots,n$ and $j=1,2,\cdots,m$, where $W[i,j]$ is the element in the $i$th row and the $j$th column of matrix $W$ and $b[i]$ is the $i$th element of vertex $b$. Consequently, we build a $(K+1)$-layer HomoMLP $\mathcal{A}'$ with ReLU as activation functions and with width $d_m\le \max(2n,n+5)$ such that,
	\begin{eqnarray}
	\int_{\mathbb{X}}|f(x)-F_{\mathcal{A}'}(x)|\dif x = \int_{\mathbb{X}}|f(x)-G_\mathcal{A}(x)|\dif x < \epsilon.
	\qed\nonumber
	\end{eqnarray}

	\section{Method}
	\label{sec:method}
	
	We present more details about our iterative module in Section~\ref{sec:iterative-module}, such as the memory-efficient implementation in Section~\ref{sec:iterative-memory-efficient},  the node-wise iterative module to support unconnected graphs in Section~\ref{sec:iterative-node-wise}, and the decaying confidence mechanism to achieve much larger iteration numbers during inference in practice in Section~\ref{sec:iterative-decay}. 
	We describe how to formulate homogeneous neural network modules in Section~\ref{sec:homogeneous-prior}. 
	We show the formulation of PathGNN layers in detail in Section~\ref{appsec:pathgnn}. There are three variants of PathGNN, each of which corresponds to different degrees of flexibilities of approximating functions. At last, we present the random initialization technique for solving the component counting problem using GNNs and discuss its motivations in Section~\ref{sec:random-init}.
	
	\subsection{Iterative module}
	\label{sec:iterative-module}
	
	We propose IterGNN to break the limitation of fixed-depth graph neural networks so that models can generalize to graphs of arbitrary scales. The core of IterGNN is a differentiable iterative module as described in the main body. We present more details about its formulations and implementations in this subsection. Its representation powers are analyzed in Section~\ref{sec:theory-iterative}.
	
	\subsubsection{Node-wise iterative module}
	\label{sec:iterative-node-wise}
	
	In the main body, we assume all nodes within the same graph share the same scale, so that we predict a single confidence score for the whole graph at each time step while building IterGNNs. However, for problems like connected component counting, the graphs can have multiple components of different scales. We can then utilize node-wise IterGNNs to achieve better performance. It is equivalent to apply our iteration module as described in the main body to each node (instead of to each graph). The models can thus learn to iterate for different times for nodes/components of different scales.
	
	In detail, we also set the iteration body function $f$ and the stopping criterion function $g$ as GNNs. The body function $f$ still update the node features iteratively $\{\vec{h}_v^{(k+1)}:v\in V\}=\text{GNN}(G, \{\vec{h}_v^{(k)}:v\in V\}, \{\vec{h}_e:e\in E\})$. On the other hand, We build the termination probability module $g$ by node-wise embedding modules and node-wise prediction modules. Typically, we apply the same MLP to features of each node to predict the confidence score of each node $c_v^k=\text{sigmoid}(\text{MLP}(\vec{h}_v^{(k)}))$ at time step $k$. We similarly take the expectation of node features as the output, however with different distributions for different nodes: $\vec h_v = \sum_{k=1}^{\infty}c^k_v\prod_{i=1}^{k-1}(1-c_v^i)\vec{h}_v^k$. 
	
	\subsubsection{Decaying confidence mechanism}
	\label{sec:iterative-decay}

	Although the vanilla IterGNN, as described in the main body, theoretically supports infinite iteration numbers, models can hardly generalize to much larger iteration numbers during inference in practice. In detail, we utilize the sigmoid function to ensure that confidence scores are between 0 and 1. However, the sigmoid function can't predict zero confidence scores to continue iterations forever. Alternatively, the models will predict small confidence scores when they are not confident enough to terminate at the current time step. As a result, the models will still work well given the IID assumption, but can't generalize well when much larger iteration numbers are needed than those met during training. For example, while solving the shortest path problem, 0.05 is a sufficiently small confidence score during training, because no iteration number larger than 30 is necessary and $0.95^{30}$ is still quite larger than 0. However, such models can not generalize to graphs with node numbers larger than 300, since $0.95^{300} \rightarrow 0$ and the models will terminate before time step 300 in any case. During our preliminary experiments, the vanilla IterGNN can not iterate for more than 100 times.
	
	The key challenge is the difference between iteration numbers during training and inference. We then introduce a simple decaying mechanism to achieve larger iteration numbers during inference. The improved algorithm is shown in Algorithm \ref{alg:iteration_decay}.
	The termination probabilities will manually decay/decrease by $\lambda$ at each time step. The final formulation of IterGNN can then generalize to iterate for 2500 times during inference in our experiments.

	We compare the choices of decaying ratios as $0.99, 0.999, 0.9999$ in our preliminary experiments and fix it to $0.9999$ afterwards in all experiments.
	
	\begin{algorithm}
		\caption{IterGNN with decay. $g$ is the stopping criterion and $f$ is the iteration body}
		\label{alg:iteration_decay}
		\begin{algorithmic}
			\STATE \textbf{Input: } initial feature $x$; stop threshold $\epsilon$; decay constant $\lambda$;
			\STATE $k\gets 1$
			\STATE $h^0\gets x$
			\WHILE{$\lambda^{k}\prod_{i=1}^{k-1}(1-c^i)>\epsilon$}
			\STATE $h^k\gets f(h^{k-1})$
			\STATE $c^k\gets g(h^k)$
			\STATE $k\gets k+1$
			\ENDWHILE
			\STATE  {\bfseries return }$h=\lambda^k\sum_{j=1}^{k}c^jh^j\prod_{i=1}^{j-1}(1-c^i)$
		\end{algorithmic}
	\end{algorithm}

	\subsubsection{Efficient train and inference by IterGNNs}
	\label{sec:iterative-memory-efficient}
	
	The advantages of IterGNN are not only limited to improving the generalizability w.r.t. scales. For example, 
	IterGNN also promotes the standard generalizability because of its better algorithm alignment~\cite{xu2019can} to iterative algorithms.
	In this subsection, we show that IterGNN moreover improves efficiencies for both training and inference. Briefly speaking, the improved generalizability w.r.t. scales enables training on smaller graphs while still achieving satisfied performance on larger graphs. The cost of computations and the cost of memories are therefore decreased on those smaller graphs during training. During inference, we implement a memory-efficient algorithm by expressing the same logic with different formulas. 
	
	In detail, training GNNs take memories whose sizes scale at least linearly (in general quadratically) with respect to the graph sizes. In our preliminary experiments, 11GB GPU memories are not enough to train 30-layer GNNs on graphs of node numbers larger than 100 when the batch size is 32. It is therefore either infeasible or super-inefficient to train 500-layer GNNs directly on graphs with 1000 nodes, to meet the IID assumptions. We couldn't even fit such models within 32GB CPU memories for training. On the other hand, with the help of IterGNN, fewer iterations and smaller graphs are needed for training to achieve satisfying performance on larger graphs.
	
	During inference, thanks to the equivalent formulations of IterGNNs as depicted in Algorithm~\ref{alg:iteration_inference}, we don't need to store the node features at all times steps until the final output of our iterative module, as done in Algorithm~\ref{alg:iteration_decay}. In practice, we can calculate the final output step by step as follows

	\begin{algorithm}
		\caption{IterGNN's efficient inference}
		\label{alg:iteration_inference}
		\begin{algorithmic}
			\STATE \textbf{Input:} initial feature $x$; stop threshold $\epsilon$; decay constant $\lambda$;
			\STATE $k\gets1;~h^0\gets x;~\bar{c} \gets 1;~\tilde{h} \gets \Vec{0};$
			\WHILE{$\bar{c} > \epsilon$}
			\STATE $h^k \gets f(h^{k-1});~c^k\gets g(h^k);$
			\STATE $\tilde{h} \gets \lambda\tilde{h}+\bar{c}c^kh^k$
			\STATE $\bar{c} \gets \lambda(1-c^k)\bar{c};$
			\STATE $k\gets k+1$
			\ENDWHILE
			\STATE  {\bfseries return }$h=\tilde{h}$
		\end{algorithmic}
	\end{algorithm}
	
	Note that this memory-efficient algorithm is only applicable during inference, since the node features at each time step must be stored during training to calculate the gradients during backward passes.
	
	\subsection{Path Graph Neural Networks}
	\label{appsec:pathgnn}
	
	As stated in~\cite{battaglia2018relational}, the performance of GNNs, especially their generalizability and zero-shot transferability, is largely influenced by the relational inductive biases. Xu~\cite{xu2019can} further formalized the relational inductive biases as sample efficiencies from algorithm alignments. For solving path-related graph problems such as shortest path, a classical algorithm is the Bellman-Fold algorithm. Therefore, to achieve more effective and generalizable solvers for path-related graph problems, we design several specializations of GN blocks, as described in~\cite{battaglia2018relational} and in Section~\ref{sec:gn-block}, by exploiting the inductive biases of the Bellman-Fold algorithms. The notations are also presented in Section~\ref{sec:gn-block}.
	
	Our first observation of the Bellman-Fold algorithm is that it directly utilizes the input attributes such as the edge weights and the source/target node identifications at each iteration. We further observe that the input graph attributes of classical graph-related problems are all informative, well defined and also well represented as discrete one-hot encodings or simple real numbers (e.g. edge weights). Therefore, we directly concatenate the input node attributes with the hidden node attributes as the new node attributes before fed into our Graph Networks (GN) blocks. The models then don't need to extract and later embed the input graph attributes into the hidden representations in each GN block. 
	
	\begin{algorithm}[tb]
		\caption{The Bellman-Fold algorithm}
		\label{alg:bellman-fold}
		\begin{algorithmic}
			\STATE {\bfseries Input:} node attributes $V=\{\textbf{v}_i,i=1,2,\cdots,N_v\}$, edge attributes $E=\{(w_k,s_k,r_k),k=1,2,\cdots,N_e\}$, and the source node $source$.
			\STATE {\bfseries Output:} The shortest path length from the source node to others, $distance=\{dist_i,i=1,2,\cdots,N_v\}$.\\~\\
			\STATE Initialize the intermediate node attributes $V'=\{dist_i,i=1,2,\cdots,N_v\}$ as $dist_i=\begin{cases}
			0 &if.~i=source\\
			\infty & o.w.
			\end{cases}$.
			\FOR{i=1 \textbf{to} $N_v-1$}
			\FOR{$(w_k,s_k,r_k)$ \textbf{in} $E$}
			\STATE $\textbf{e}'_k=dist_{s_k}+w_k$ ~~~$\triangleleft$ edge message module
			\ENDFOR
			\FOR {j=1 \textbf{to} $N_v$}
			\STATE $\bar{\textbf{e}}_j=\min(\{\textbf{e}'_k:r_k=j\})$ ~~~$\triangleleft$ aggregation module
			\STATE $dist_{j}=\min(dist_j,\bar{\textbf{e}}_j)$ ~~~$\triangleleft$ update module
			\ENDFOR 
			\ENDFOR
		\end{algorithmic}
	\end{algorithm}
	
	Our second observation is that Bellman-Fold algorithm can be perfectly represented by Graph Networks as stated in Algorithm \ref{alg:bellman-fold}. Typically, for each iteration of the Bellman-Fold algorithm, the message module sums up the sender node's attributes (i.e. distance) with the edge weights, the node-level aggregation module then selects the minimum of all edge messages, and finally the attributes of the central node are updated if the new message (/distance) is smaller. The other modules of GN blocks are either identity functions or irrelevant. Therefore, to imitate the Bellman-Fold algorithm by GN blocks, we utilize max pooling for both aggregation and update modules to imitate the min poolings in the Bellman-Fold algorithm. The edge message module is MLP,  similarly to most GNN variants. The resulting module is then equal to replacing the MPNN \cite{gilmer2017neural}'s aggregation and update module with max poolings. Therefore, we call it MPNN-Max as in \cite{velivckovic2019neural}. The concrete formulas are as follows
	\begin{eqnarray*}
		\textbf{e}'_k&=&MLP(\textbf{v}_{s_k},\textbf{v}_{r_k},\textbf{e}_k)\\
		\bar{\textbf{e}}_j&=&\max(\{\textbf{e}'_k:r_k=j\})\\
		\textbf{v}'_{j} & = & \max(\textbf{v}_j,\bar{\textbf{e}}_j)
	\end{eqnarray*}
	
	Moreover, we notice that the Bellman-Ford algorithm is only designed for solving the shortest path problem. Many path-related graph problems can not be solved by it. Therefore, we further relax the max pooling to attentional poolings to increase the models' flexibility while still maintaining the ability to approximate min pooling in a sample efficient way. Typically, we propose the PathGNN by replacing the aggregation module with attentional pooling. The detailed algorithm is stated in Algorithm \ref{alg:pathgnn}, where the global attributes are omitted due to their irrelevance.
	
	\begin{algorithm}[tb]
		\caption{One step of PathGNN}
		\label{alg:pathgnn}
		\begin{algorithmic}
			\STATE {\bfseries Input:} graph $G=(V,E)$
			\STATE {\bfseries Output:} updated graph $G'=(V',E)$\\~\\
			
			\FOR{$(w_k,s_k,r_k)$ \textbf{in} $E$}
			\STATE $\tilde{\textbf{e}}_k=MLP(\textbf{v}_{s_k},\textbf{v}_{r_k},\textbf{e}_k)$
			\STATE $score_k=MLP'(\textbf{v}_{s_k},\textbf{v}_{r_k},\textbf{e}_k)$
			\STATE $\textbf{e}'_k=(score_k, \tilde{\textbf{e}}_k)$ ~~~$\triangleleft$ edge message module
			\ENDFOR
			\FOR {j=1 \textbf{to} $N_v$}
			\STATE $\bar{\textbf{e}}_j=\textrm{attention}(\{\textbf{e}'_k:r_k=j\})$ ~~~$\triangleleft$ aggregation module
			\STATE $\textbf{v}'_{j}=\max(\textbf{v}_j,\bar{\textbf{e}}_j)$ ~~~$\triangleleft$ update module
			\ENDFOR 
		\end{algorithmic}
	\end{algorithm}
	
	\begin{algorithm}[tb]
		\caption{Attention Pooling in GNNs}
		\label{alg:attention}
		\begin{algorithmic}
			\STATE {\bfseries Input:} set of messages $\{\textbf{e}'_k=(score_k,\tilde{\textbf{e}}_k)\}$ 
			\STATE {\bfseries Output:} aggregated messages $\bar{\textbf{e}}_j$\\~\\
			
			\STATE$\alpha = softmax(score)$
			\STATE $\bar{\textbf{e}}_j=\sum_k\alpha_k\tilde{\textbf{e}}_k$
		\end{algorithmic}
	\end{algorithm}
	
	Another variant of PathGNN is also designed by exploiting a less significant inductive bias of the Bellman-Fold algorithm. Specifically, we observe that only the sender node's attributes and the edge attributes are useful in the message module while approximating the Bellman-Fold algorithm. Therefore, we only feed those attributes into the message module of our new PathGNN variant, PathGNN-sim. The detailed algorithm is stated in Algorithm \ref{alg:pathgnn-sim}.
	
	\begin{algorithm}[tb]
		\caption{One step of PathGNN-sim}
		\label{alg:pathgnn-sim}
		\begin{algorithmic}
			\STATE {\bfseries Input:} graph $G=(V,E)$
			\STATE {\bfseries Output:} updated graph $G'=(V',E)$\\~\\
			
			\FOR{$(w_k,s_k,r_k)$ \textbf{in} $E$}
			\STATE $\tilde{\textbf{e}}_k=MLP(\textbf{v}_{s_k},\textbf{e}_k)$
			\STATE $score_k=MLP'(\textbf{v}_{s_k},\textbf{v}_{r_k},\textbf{e}_k)$
			\STATE $\textbf{e}'_k=(score_k, \tilde{\textbf{e}}_k)$ ~~~$\triangleleft$ edge message module
			\ENDFOR
			\FOR {j=1 \textbf{to} $N_v$}
			\STATE $\bar{\textbf{e}}_j=\textrm{attention}(\{\textbf{e}'_k:r_k=j\})$ ~~~$\triangleleft$ aggregation module
			\STATE $\textbf{v}'_{j}=\max(\textbf{v}_j,\bar{\textbf{e}}_j)$ ~~~$\triangleleft$ update module
			\ENDFOR 
		\end{algorithmic}
	\end{algorithm}
	
	In summary, we introduce Path Graph Neural Networks (PathGNN) to improve the generalizability of GNNs for distance related problems by improving the algorithm alignment~\cite{xu2019can}. It is a specially designed GNN layer that imitates one iteration of the classical Bellman-Ford algorithm. There are three variants of PathGNN, i.e. MPNN-Max, PathGNN, and PathGNN-sim, each of which corresponds to different degrees of flexibilities.
	In our experiments, they perform much better than GCN and GAT for all path-related tasks regarding the generalizability, as stated in Section 5 in the main body.

	\subsection{Homogeneous prior}
	\label{sec:homogeneous-prior}
	
	As described in the main body, the approach to build HomoGNN is simple: remove all the bias terms in the multi-layer perceptron (MLP) used by ordinary GNNs, so that all affine transformations degenerate to linear transformations. Additionally, only activation functions that are homogeneous are allowed to be used.
	Applying this approach to GN-blocks, we have the homogeneous GN blocks. The HomoMLP is defined as MLPs without biases and with homogeneous functions as the activation functions. Note that ReLU and Leaky ReLU are both homogeneous functions. The sum/max/mean poolings are also homogeneous functions.
	
	The only non-homogeneous pooling module that is widely used in GNNs is the attentional pooling module~\cite{girdhar2017attentional, vaswani2017attention}. We also utilizes it as a flexible aggregation module in PathGNN as described in Section~\ref{appsec:pathgnn}. In this subsection, we present another simple approach, which is generally similar to the previous one, to design the homogeneous attentional poolings: replace MLPs with HomoMLPs and apply one normalization layer before softmax. In detail, most attentional poolings have similar architecture to the one in PathGNNs as presented in Algorithm~\ref{alg:attention}. The attentional pooling modules output the weighted summation of updated features $\tilde{\textbf{e}}_k$, where the weights $\alpha_k$ are probabilities calculated by softmax based on the scores $score_k$. The updates features $\tilde{\textbf{e}}_k$ and scores $score_k$ are both calculated by applying some MLPs on the input features, as shown in Algorithm\ref{alg:pathgnn}. We want to design attentional poolings that are homogeneous functions over the input features. The approach is as follows: We change all MLPs to HomoMLPs and replace softmax with a scale-invariant version of softmax. In this case, the weights $\alpha_k$ will not change with respect to the scales of input features. The magnitudes of the final output of attentional pooling modules $\bar{\textbf{e}}$, which is the weighted summation of updated features $\tilde{\textbf{e}}_k$, will then scale linearly with respect to the magnitudes of updates features as well as the magnitudes of input features. Therefore, the resulting attentional module is a homogeneous function. To design a scale-invariant softmax, we simply adopt one normalization layer before softmax so that the effect of scales is eliminated. Typically, we build our scale-invariant softmax as 
	\begin{eqnarray}
	\text{scale-invariant-softmax}(score) = \text{softmax}\left( \frac{score}{\max score-\min score}\right) 
	\end{eqnarray}
	
	No division will be performed if $\max score=\min score$. Together with the bias-invariance property of softmax (Proposition 2 in~\cite{martins2016softmax}), we have for all $\lambda>0$,
	\begin{eqnarray}
	\text{scale-invariant-softmax}(\lambda\cdot score) = \text{scale-invariant-softmax}(score)
	\end{eqnarray}
	
	Similar intuitions can be applied to design attentional poolings with activation functions other than softmax, such as sparsemax~\cite{martins2016softmax}.
	
	\subsection{Random Initialization}
	\label{sec:random-init}

	\begin{figure*}
		\centering 
		\includegraphics[width=\textwidth]{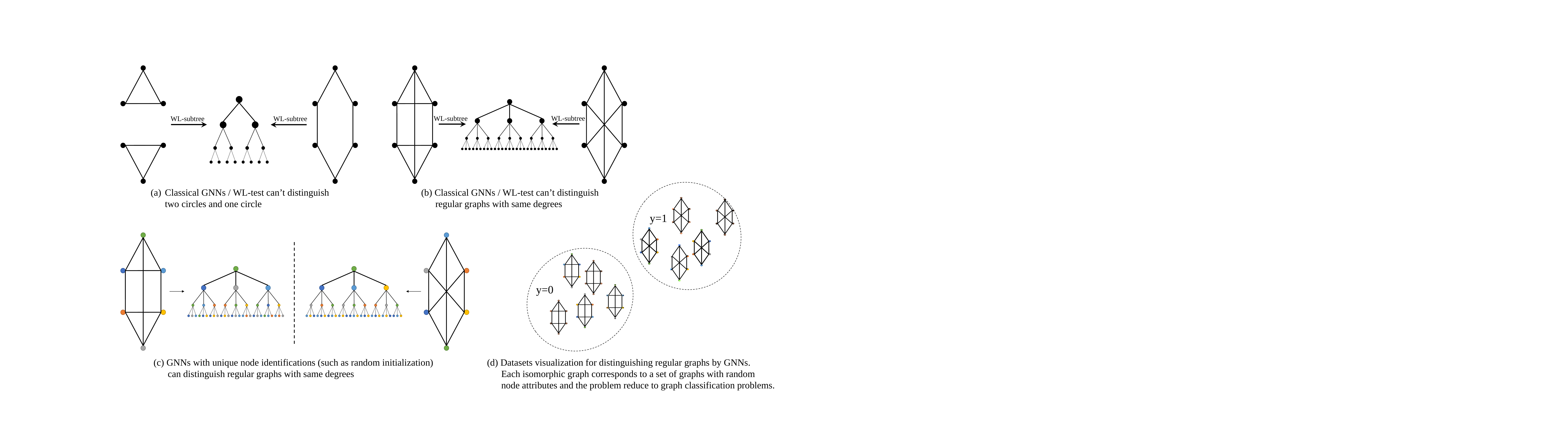}
		\caption{Random initialization improves the representation power of GNNs by distinguishing different nodes. (a) and (b) illustrate that classical GNNs which are as powerful as WL-test for graph isomorphic problems can not distinguish regular graphs with same degrees. Typically, (a) shows an example of two graphs with different component numbers and (b) shows an example of two graphs with different diameters. Therefore, classical GNNs with constant initialization are not powerful enough to solve many graph-related problems such as component counting and graph diameters. (c) illustrates that GNNs with random initialization can distinguish regular graphs since each node is assigned with a unique identification. (d) further demonstrates the details about datasets for training and evaluating GNNs with random initialization. Specifically, GNNs can map each isomorphic graph into a set of vectors in the graph-level embedding space and then the graph-isomorphic problem reduces to the classical classification problem.}
		\label{fig:random-init}
	\end{figure*}

	As analyzed in previous works~\cite{xu2018how, morris2019weisfeiler}, standard GNNs are at most as powerful as the Weisfeiler-Leman test (WL-test)~\cite{weisfeiler1968reduction} for distinguishing non-isomorphic graphs. Standard GNNs are then theoretically short of representation powers for solving graph problems such as component counting and graph radius/diameters, as illustrated in Figure \ref{fig:random-init}. WL-test can't distinguish the graph as one circle with 6 nodes and the graph as two circles each with 3 nodes.
	
	However, previous works all assumed that the node attributes were initialized as constant while analyzing GNNs' representation powers. With the constant initialization, nodes with the same subtree patterns are not distinguishable similarly to the WL-test as shown in Figure \ref{fig:random-init}. Nevertheless, they didn't fully utilize the representation powers of neural networks. Instead of utilizing the constant initialization techniques, the node attributes could be initialized as random numbers so that each node has its own unique identification. Subtrees that are of the same pattern but are composed with different nodes are therefore distinguishable. In this case, each graph structure $E$ is mapped into a set of input graphs $G_{random}=\{V_{random},E\}$ where $V_{random} = \{\textbf{v}_i=rand(), i=1,2,\cdots,N_v\}$. And the graph-isomorphic problem is then formulated as distinguishing sets of graphs. As illustrated in Figure \ref{fig:random-init}, and also as verified in Section \ref{sec:exp-random-init}, GNNs with random initialization can distinguish non-isomorphic regular graphs which are unable to be distinguished by WL-test. Therefore, random initialization can improve the representation powers of GNNs. \cite{murphy2019relational, anonymous2020coloring} also formulated the improvement of representation powers by random initialization while from another perspective, specifically towards efficient universal approximators of permutation invariant/equivalent functions.

	\section{Experiment Setups}
	\label{sec:exp-setups}
	
	We present the detailed experimental setups in this subsection. We first present details of three graph theory problems, i.e. the shortest path problem in Section~\ref{sec:exp-setup-shortest-path}, the component counting problem in Section~\ref{sec:exp-setup-comp-cnt}, and the traveler salesman problem in Section~\ref{sec:exp-setup-tsp}. We then list the experimental setups of three graph-based reasoning tasks, i.e. the physical simulation in Section~\ref{sec:exp-setup-phy-sim}, the image-based navigation in Section~\ref{sec:exp-setup-img-navi}, and the symbolic Pacman in Section~\ref{sec:exp-setup-pacman}. We also describe the experimental setups of graph classification in Section~\ref{sec:exp-setup-graph-classification}. The descriptions of models are stated in Section~\ref{sec:exp.models} and the training details are presented in Section~\ref{sec:training-details}.
	
	\subsection{Graph theory problems}
	
	We evaluate our proposals on three classical graph theory problems, i.e. shortest path, component counting, and the Traveling Salesman Problem (TSP). We present the properties of graph generators in Section~\ref{sec:graph-generators' properties}, the setups for the shortest path problem in Section~\ref{sec:exp-setup-shortest-path}, the setups for the component counting in Section~\ref{sec:exp-setup-comp-cnt}, and the setups for the TSP problem in Section~\ref{sec:exp-setup-tsp}.
	
	\subsubsection{Properties of Graph Generators}
	\label{sec:graph-generators' properties}
	
	How to sample graphs turns to be intricate in our exploration to deeply investigate the generalizability power. Four generators are adopted in our experiments:
	\begin{itemize}
		\item The Erdos-Renyi model \cite{erd6s1959random}, $G(n,p)$, generates graphs with $n$ nodes and each pair of nodes is connected with probability $p$. 
		\item The KNN model, $\text{KNN}(n, d, k)$, first generates $n$ nodes whose positions are uniformly sampled from a $d$-dimensional unit cube. The nodes are then connected to their $k$ nearest neighbors. The edge directions are from the center node to its neighborhoods.
		\item The planar model, $\text{PL}(n, d)$, first generates $n$ nodes whose positions are uniformly sampled from a $d$-dimensional unit cube. Delaunary triangulations are then computed \cite{barber1996quickhull} and nodes in the same triangulation are connected to each other.
		\item The lobster model, $\text{Lob}(n,p_1,p_2)$, first generates a line with $n$ nodes. $n_1\sim \mathcal{B}(n, p_1)$ nodes are then generated as first-level leaf nodes, where $\mathcal{B}$ denotes the binomial distribution. Each leaf node uniformly selects one node in the line as the parent. Afterwards, $n_2\sim \mathcal{B}(n_1,p_2)$ are generated as second-level leaf nodes. Each second-level leaf node also uniformly selects one first-level leaf node as the parent. The parents and children are connected to each other and the graph is therefore undirected.
	\end{itemize}
	
\begin{figure*}[]
		\centering 
		\includegraphics[width=\textwidth]{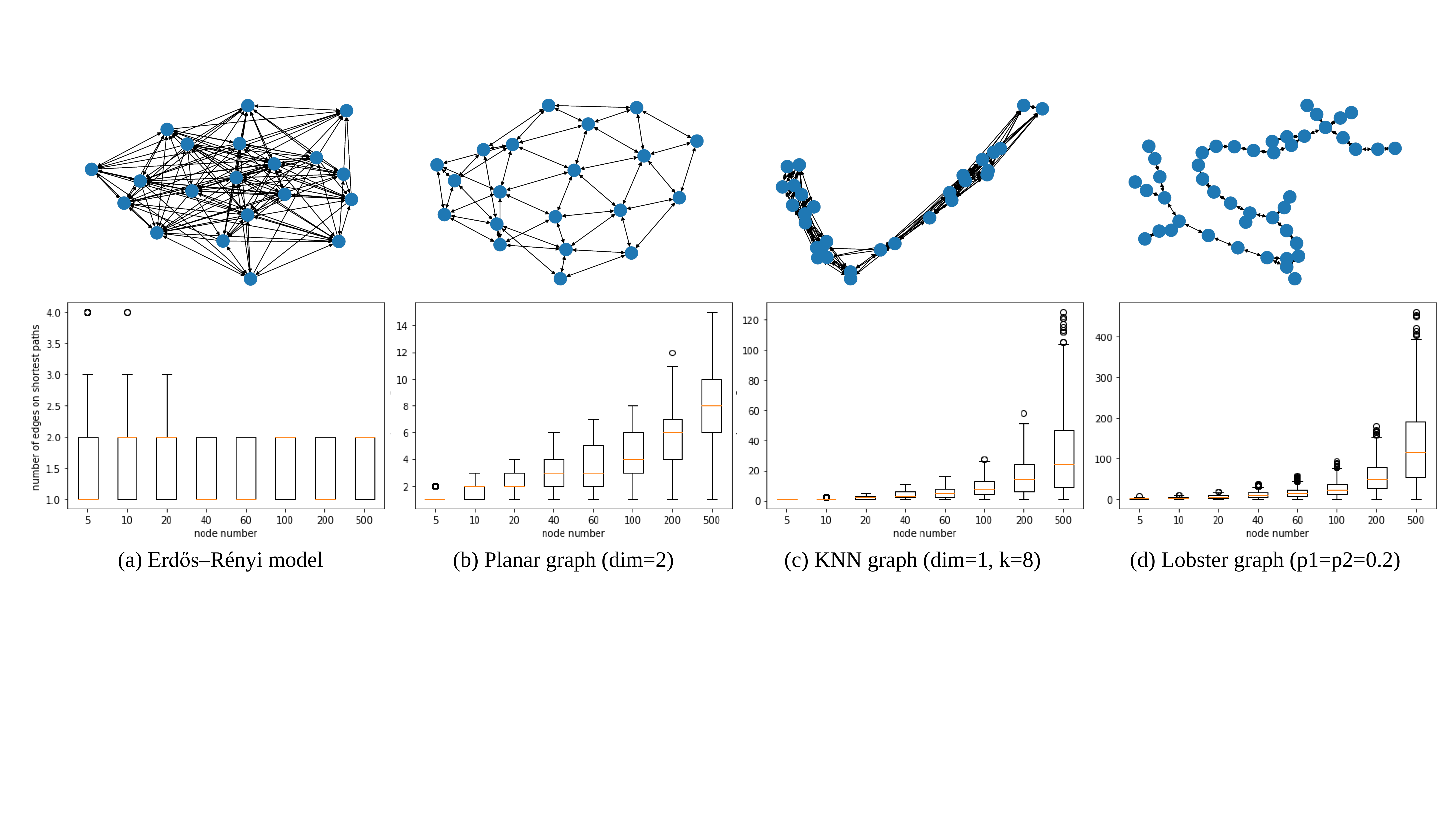}
		\caption{Properties of different random graph generators. The upper row illustrates graph samples generated by the corresponding generator. The lower row demonstrates the relationships between the graph sizes (i.e. node numbers) and the distributions of random node pairs' distances. Typically, for each generator and each graph size, 1000 sample graphs are generated by the corresponding generator for estimating the distributions. Box plots are utilized for visualizing the distributions.}
		\label{fig:random-graphs}
	\end{figure*}

	For the Erdos-Renyi model, we assign $p$ equal to 0.5 so that all graphs of $n$ nodes can be generated with equal probabilities. However, the expectation of graph diameters decreases dramatically as graph sizes increase for such model. As illustrated in Figure \ref{fig:random-graphs}, the graph diameters are just 2 with high probability when the node number is larger than 50.
	
	The other three graph generators are therefore designed to generate graphs with larger diameters for better evaluation of models' generalizabilities w.r.t. scales. We manually select their hyper-parameters to efficiently generate graphs of diameters as large as possible. Specifically, we set $d=1$ and $k=8$ for the KNN model, $d=2$ for the planar model, $p_1=0.2$ and $p_2=0.2$ for the lobster model. Their properties are illustrated in Figure \ref{fig:random-graphs}. Note that the distances are positively related to the graph sizes for all three graph generators. Moreover, for the lobster model, the distances increase almost linearly with respect to the graph sizes.

	\subsubsection{Shortest Path}
	\label{sec:exp-setup-shortest-path}
	
	In this problem, given a source node and a target node, the model needs to predict the length of the shortest path between them. The edge weights are positive and uniformly sampled. We consider both unweighted graphs and weighted graphs (edge weights uniformly sampled between 0.5 and 1.5). Groundtruth is calculated by Dijkstra's algorithm.
	
	We utilize a three-dimensional one-hot representation to encode the location of the source and target nodes ($100$ for source, $010$ for destination, and $001$ for other nodes). Edge weights are encoded as the edge attributes. Two metrics are used to measure the performance of GNNs. The relative loss is first applied to measure the performance of predicting shortest path lengths. To further examine the models' ability in tracing the shortest path, we implement a simple post-processing method leveraging the noisy approximation of path lengths (described later). We define the relative loss as $|l-l_{pred}|/l$ and the success rate of identifying the shortest path after post-processing as $\mathbbm{1}(l=l_{post\text{-}pred})$, where $l$ is the true shortest path length, $l_{pred}$ is the predicted length by GNNs, and $l_{post\text{-}pred}$ is the length of the predicted path after post-processing.
	
	\paragraph{Post-Processing}
	\label{sec:post-processing}
	
	After training, our model can predict the shortest path length from source to target nodes. The post-processing method is then applied to find the shortest path based on the learned models. Specifically, the post-processing method predicts the shortest path $$p=[p_1,p_2,\cdots,p_n]=\textrm{post-processing}(G;GNN)$$
	given the input graph $G=(V_{i,j},E)$, as defined in Section 5.1.1., and the noisy shortest path length predicting model $GNN$, where $i,j$ are the indexes of source and target nodes, $V_{i,j}$ is the node attributes with one-hot encodings, and $p_k$ denotes the index of the $k$th node on the shortest path. 
	
	The post-processing algorithm is stated in Algorithm.\ref{alg:path-post}. Concretely, we first denote the shortest path length between any two nodes $i'$ and $j'$ predicted by the trained model $GNN$ as $dist_{i',j'} = GNN(G_{i',j'}) = GNN((V_{i',j'},E))$, where $V_{i',j'}$ represents the one-hot node attributes when nodes $i'$ and $j'$ are the source and target nodes. For further convenience, we also defined $w_{i',j'}$ as the weight of edge connecting node $i'$ and node $j'$ ($w_{i',j'}=\infty$ if node $i'$ and node $j'$ are not connected by an edge.). Then, the post-processing algorithm sequentially predicts the next node $p_{k+1}$ of node $p_k$ by minimizing the difference between the predicted shortest path length approximated by GNNs $dist_{p_k,j}$ and the length of shortest path predicted by the post-processing method $dist_{p_{k+1},j}+w_{p_k,p_{k+1}}$. To further reduce the effect of models' noises, another constrain is added as $dist_{p_{k+1},j}+w_{p_k,p_{k+1}}\le dist_{p_k,j}$ so that the method will always convergence.

	\begin{algorithm}[tb]
		\caption{Post-processing to predict the shortest path}
		\label{alg:path-post}
		\begin{algorithmic}
			\STATE {\bfseries Input:} graph $G=(V_{i,j},E)$; source node index $i$; target node index $j$; trained models $GNN$ that predict the shortest path length from nodes $i'$ to node $j'$ as $dist_{i',j'} = GNN((V_{i',j'},E))$
			\STATE {\bfseries Output:} shortest path $p=[p_1,p_2,\cdots,p_n]$\\~
			\STATE Initialize $p_1=i, k=1$.
			\WHILE{$p_k\neq j$}
			\IF{$|\{l:dist_{l,j}+w_{p_k,l}\le dist_{p_k,j}\}| > 0 $}
			\STATE $p_{k+1} = \arg\min_{l:dist_{l,j}+w_{p_k,l}\le dist_{p_k,j}}|dist_{l,j}+w_{p_k,l}-dist_{p_k,j}|$.
			\ELSE \STATE return $p=[]$
			\ENDIF
			\STATE $k=k+1$.
			\ENDWHILE
			\STATE \text{return} p
		\end{algorithmic}
	\end{algorithm}

	\subsubsection{Component Counting}
	\label{sec:exp-setup-comp-cnt}
	In the component counting problem, the model counts the number of connected components of an undirected graph. To generate a graph with multiple components, we first sample a random integer $m$ between $1$ to $6$ as the number of components, and then divide the nodes into $m$ parts. In detail, for $n$ nodes and $m$ components, we first uniformly sample $m-1$ positions from $1$ to $n-1$ and then divide the $n$ nodes into $m$ parts by the $m-1$ positions.
	We then connect nodes in each component using the random graph generator defined in Section~\ref{sec:graph-generators' properties}, e.g., the Erdos-Renyi graph and the lobster graph. The metric is the accuracy of correct counting.
	We initialize the node attributes by random values$\in[0,1)$, so that GNNs are powerful enough solve the component counting problem, as discussed in Section~\ref{sec:random-init}.
	
	\subsubsection{Traveler Salesman Problem (TSP)}
	\label{sec:exp-setup-tsp}
	In the Euclidean travelling salesman problem (TSP), there are several 2D points located in the Euclidean plane, and the model generates a shortest route to visit each point. The graph is complete. The weight of an edge is the Euclidean distance between the two ends.
	Points $\{(x_i,y_i)\}$ are uniformly sampled from $\{x,y\in \mathbb{Z}:1\le x,y\le 1000\}$.
	We use the standard solver for TSP, Concorde~\cite{Concorde}, to calculate the ground truth. The node attributes are the 2D coordinates of each node. We use relative loss defined the same as the shortest path problem to evaluate the networks.
	
	\subsection{Graph-based reasoning tasks}
	
	We further evaluate the benefits of our proposals using three graph-based reasoning tasks. We describe the setups of physical simulation in Section~\ref{sec:exp-setup-phy-sim}, the setups of symbolic Pacman in Section~\ref{sec:exp-setup-pacman}, and the setups of image-based navigation in Section~\ref{sec:exp-setup-img-navi}.
	
	\begin{figure}
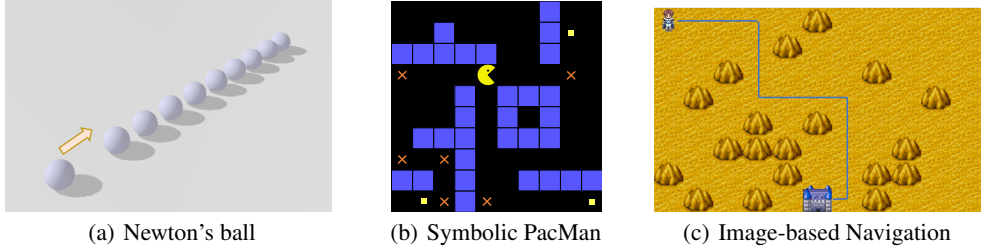

		\centering
		\subfigure[Newton's ball]{\includegraphics[width=0.31\textwidth]{figure/newtonball.png} \label{appfig:newton}}%
		\qquad
		\subfigure[Symbolic PacMan]{\includegraphics[width=0.2\textwidth]{figure/pacman.pdf}\label{appfig:symbolic_pacman}}%
		\qquad
		\subfigure[Image-based Navigation]{\includegraphics[width=0.31\textwidth]{figure/image-based-navi.png}\label{appfig:image-navi}}
		\caption{Figure (a) shows a set of Newton's balls in the physical simulator. The yellow arrow shows the moving direction of the first ball. Figure (b) is a scene in our symbolic PacMan environment. Figure (c) illustrates our image-based navigation task in a RPG-game environment.}
		\label{appfig:general-task}
	\end{figure}{}
	
	\subsubsection{Physical Simulation}
	\label{sec:exp-setup-phy-sim}
	
	We evaluate the generalizability of our models by predicting the moving patterns between objects in a physical simulator. We consider an environment 
	similar to \textit{Newton's cradle}, also known as 
	called \textit{Newton's ball}, as shown in Figure \ref{appfig:newton}: 
	all balls with the same mass lie on a friction-free pathway. With the ball at one end moving towards others, our model needs to predict the motion of the balls of both ends at the next time step. 
	The probability is 50\% for balls to collide.
	We represent the environment as a chain graph. The nodes of the graph stand for the balls and the edges of the graph stand for the interactions between two adjacent balls. We fix the number of balls within $[4,34)$ at the training phase, while test the networks in environments with 100 nodes.
	
	In detail, we generate samples as follows: 
	\begin{itemize}
		\item $n-1$ balls with same properties are placed as a chain in the one-dimensional space where each ball touches its neighbourhoods.
		\item A new ball moves towards the $n-1$ balls from left position $x$ with constant speed $v$.
		\item The model needs to predict each ball's position and speed after one time step.
	\end{itemize} 
	The radius of balls, $r$, is set to 0.1 and the positions are normalized so that the origin is in the middle of $n-1$ balls. The left ball's position $x$ is set so that its distance to the most left ball among the other $n-1$ balls is uniformly sampled between $0$ and $200r$. The left ball's speed $v$ is uniformly sampled between $0$ and $200r$. Note that the left ball may or may not collide with the other balls depending on its positions and weights. The probability is 50\% for balls to collide.
	
	Since the $n-2$ balls in the middle will not change their positions or speeds in any cases, we simplify the output to the positions and speeds of left and right balls. To avoid trivial solutions, we still force the models to predict positions and speeds at the node level, which means no global readout modules are allowable. 
	
	\subsubsection{Image-based Navigation}
	\label{sec:exp-setup-img-navi}
	
	We show benefits of the differentiability of a generalizable reasoning module using the image-based navigation task,
	as illustrated in Figure \ref{appfig:image-navi}. 
	The model needs to plan the shortest route from source to target on 2D images with obstacles. However, the properties of obstacles are not given as a prior and the model must discover them based on image patterns during training. 
	
	We simplify the task by defining each pixel as obstacles merely according to its own pixel values.
	Specially, we assign random heights from $[0,1)$ to pixels in 2D images. The agent can't go through pixels with heights larger than 0.8 during navigation. The cases where no path exists between the source node and the target node are abandoned.
	We represent the image by a grid graph. Each node corresponds to a pixel. The edge attributes are set to one. The node attributes are  the concatenation of pixel values and the one-hot embedding of node's categories (source/target/others). Note that, for more complex tasks, the node attributes can also include features extracted by CNNs.

	In detail, we generate samples as follows:
	\begin{itemize}
		\item $n\times n$ grid is first generated. Each node is connected to its left, right, up, and bottom neighbourhoods. (The boundary situations are omitted for simplicity)
		\item The height, $h_i$, is uniformly sampled between 0 and 1, and is then assigned to node $i$. Nodes with heights larger than 0.8 can't be visited.
		\item The source node and the target node are uniformly sampled from node pairs that have height less than 0.8 and are connected.
	\end{itemize}
	
\begin{figure}
		\centering
		\includegraphics[width=0.7\textwidth]{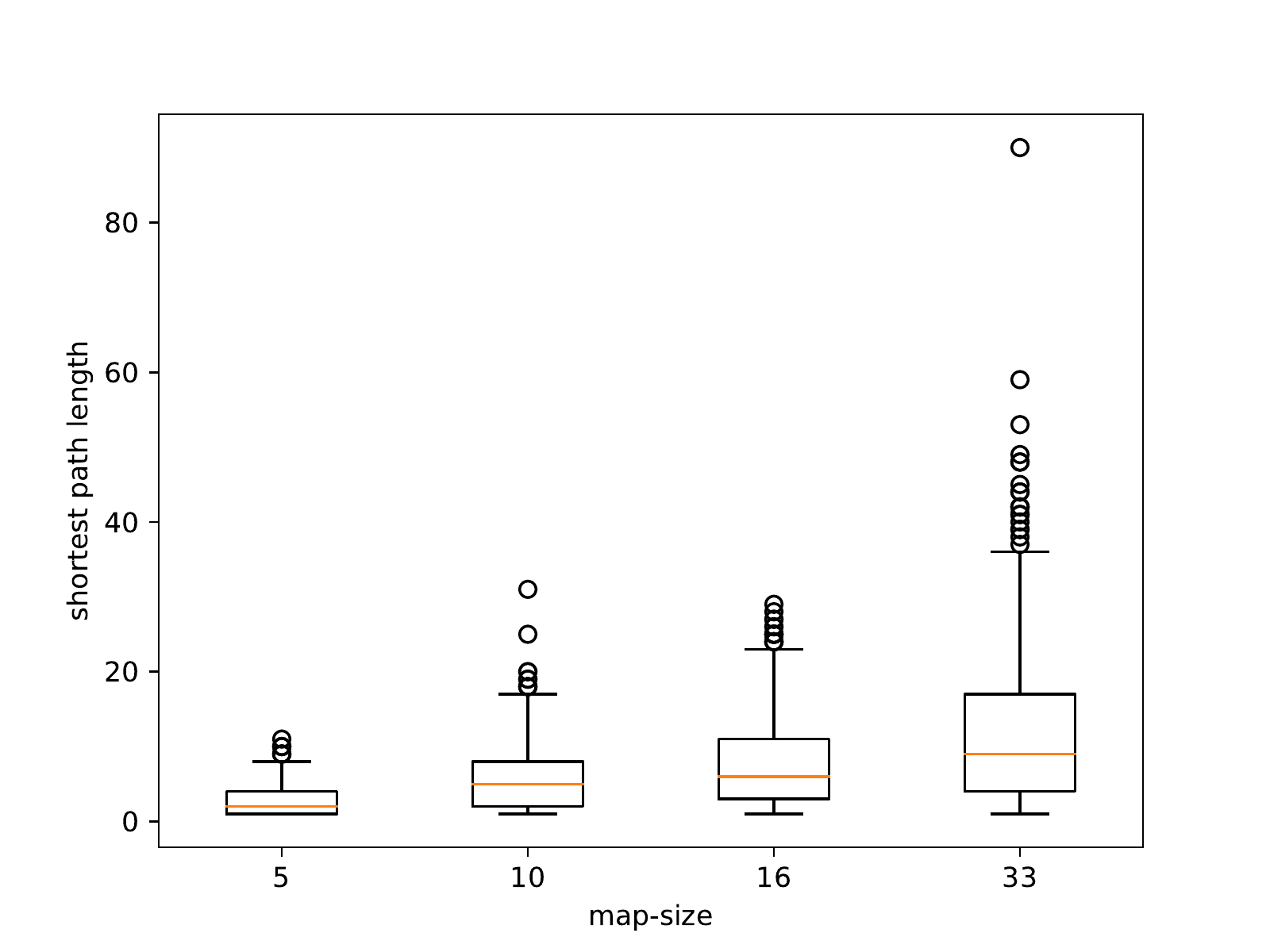}
		\caption{Illustration of the properties of the datasets for image-based navigation. Box plots are utilized to visualize the distributions of the lengths of the shortest path between two random nodes for each map-size. The shortest path lengths increase with the map sizes.}
		\label{fig:maze-property}
	\end{figure}

	The node attributes are their heights and the one-hot encodings of their categories (i.e. source, target, or others). The edge attributes are all ones. The properties of datasets are visualized in Figure \ref{fig:maze-property}.
	
	\subsubsection{Symbolic Pacman}
	\label{sec:exp-setup-pacman}
	
	To show that our iterative module can improve reinforcement learning, we construct a symbolic PacMan environment with similar rules to the PacMan in Atari ~\cite{bellemare13arcade}. 
	As shown in Figure~\ref{appfig:symbolic_pacman}, the environment contains a map with dots and walls. 
	The agent starts at one position and at each step it can move to one of neighboring grids. The agent will receive reward of $1$ when it reaches one dot and ``eats'' the dot. The discount factor is set to 0.9. 
	The agent needs to figure out a policy to quickly ``eat'' all dots while avoiding walls on the map to maximize the return.
	
	In detail, we generate random environments as follows: 
	\begin{itemize}
		\item Maze of size $16\times 16$ is first generated.
		\item $n_w$ walls of length 3 are then generated.
		\item The walls' directions are assigned randomly as vertical or horizontal.
		\item The walls' positions are uniformly sampled from all feasible ones regardless of overlappings.
		\item  $n_d$ dots are further generated with positions uniformly sampled from positions that are not occupied by walls.
		\item one agent is at last generated with positions uniformly sampled from non-occupied ones.
	\end{itemize}
	
	The model then controls the agent to navigate in the maze to eat all dots. The action space is $[$left, right, up, down$]$. The agent will not move if the action is infeasible such as colliding with the walls. The game will stop if the agent has eaten all dots or has exceeded the maximum time step, which is $16\times 16\times (n_d+1)$. A new environment will be generated afterwards. The metric is then the success rates of eating dots:
	$$\frac{\text{number of eaten dots}}{\text{number of reachable dots}}$$
	
	At each step, landmarks are placed on each corner of the shortest paths from the agent to dots. The input state is then a graph with the agent, dots, and landmarks as nodes. The node attributes are their positions plus one-hot encodings of their categories. The positions are normalized so that the agent is at the origin. The edge weights are set as the Manhattan distance between every two nodes.

	We train our models using the double DQN \cite{van2016deep} with value networks replaced by our backbones. The reward for eating each dot is 1 and no penalty for colliding with walls. The discount is set to 0.9 for each time step to encourage faster navigation. The exploration probability is 0.1 and the warm-up exploration steps are 1000. Value networks are trained every 4 time steps and updated every 200 time steps. The size of replay buffer is 10000. The batch size is 32 and the learning rate is 0.0002. Models are tested on 200 different environments and the averaged performance is reported.
	
	The network architectures are as follows: 2-layer MLP with leaky ReLU for feature embedding, GNN/CNN modules for message passing, max pooling for readout, and 1-layer FC for predicting the Q values. The hidden sizes are 64. We compare our IterGNNs with PointNet \cite{qi2017pointnet}, GCN \cite{kipf2017semi}, and CNNs. For the PointNet model, the GNN modules are identities by definition. For the GCN and CNN models, we compare the performance of their 1/3/5/7/9-layer variants and report the best of them. We also compare the choice of the kernel sizes of CNNs among 3/5/7 and report the best of them.
	
	\subsection{Graph Classification}
	\label{sec:exp-setup-graph-classification}
	Note that the abilities of models to utilize information of the long-term relationships are necessary for accurately solving most of the previous tasks and problems. Therefore, the benefits of adaptive and unbounded depths introduced by our iterative module are distinguished. In this sub-section, we show that our IterGNN can also achieve competitive performance on graph-classification benchmarks, demonstrating that our iterative module does not hurt the standard generalizability of GNNs while improving their generalizability w.r.t. graph scales. The results are presented in Section~\ref{sec:exp-graph-classification}.
	
	In detaile, we evaluate models on five small datasets, which are two social network datasets (IMDB-B and IMDB-M) and three bioinformatics datasets (MUTAG, PROTEINS and PTC). Readers are referred to \cite{xu2018how} for more descriptions of the properties of datasets. 
	We adopt the same evaluation method and metrics as previous state-of-art \cite{xu2018how}, such as 10-fold cross validation. The dataset splitting strategy and pre-processing methods are all identical to \cite{xu2018how} by directly integrating their public codes\footnote{https://github.com/weihua916/powerful-gnns}.
	
	Regarding the models, we adopt the previous state-of-art, GIN \cite{xu2018how} as GN-blocks and the JK connections \cite{xu2018representation} plus average/max pooling as the readout module. To integrate the iterative architecture, we wrap each GN-block in original backbones with the iterative module with maximum iteration number equal to 10. The tunable hyper-parameters include the number of IterGNNs, the epoch numbers, and whether or not utilizing the random initialization of node attributes.

	\subsection{Models and Baselines}
	\label{sec:exp.models}
	
	We follow the common practice of designing GNN models as presented in Section~\ref{sec:gnns}. We utilize a 2-layer MLP for node attribute embedding and use a 1-layer MLP for prediction. The max/sum poolings are adopted as readout functions to summarize information of a graph into one vector. 
	
	To build the core GNN module, we need to specify three properties of GNNs: the GNN layers, the paradigms to compose GN-blocks, and the prior, as stated in Section~\ref{sec:gnns}. In our experiments, we explore the following options for each property:
	\begin{itemize}
		\item GNN layers: PathGNN layers and two baselines that are GCN~\cite{kipf2017semi} and GAT~\cite{velickovic2018graph}.
		\item Prior: whether or not apply the homogeneous prior
		\item Paradigm to compose GN-blocks: our iterative module; the simplest paradigm that stacks multiple GNN layers sequentially; the ACT algorithm~\cite{graves2016adaptive}; and the fixed-depth shared-weights paradigm, as described in the main body.
	\end{itemize}
	
	For the homogeneous prior, we apply the prior to the node-wise embedding module, the readout module, and the final prediction module as well for most problems and tasks. However, for problems whose solutions are not homogeneous (e.g. component counting), we only apply the homogeneous prior to the core GNN module.
	
	In detail, models are specified by the choices of the previous properties. The corresponding models and their short names are as follows:
	\begin{itemize}
		\item \textit{GCN, GAT}: Models utilizing the multi-layer architecture (i.e. stacking multiple GN-blocks) with GCN \cite{kipf2017semi} or GAT \cite{velickovic2018graph} as GN-blocks. The homogeneous prior is not applied.
		\item \textit{Path / Multi-Path}: Models utilizing the multi-layer architecture and adopting PathGNN as defined in Section~\ref{appsec:pathgnn} as GN-blocks. No homogeneous prior is applied.
		\item \textit{Homo-Path / Multi-Homo-Path}: Models utilizing the multi-layer architecture and adopting PathGNN as GN-blocks. And the homogeneous prior is applied on all modules unless otherwise specified.
		\item \textit{Iter-Homo-Path / Iter-HP}: Models utilizing the iterative module as described in Section~\ref{sec:iterative-decay} and adopting PathGNN as GN-blocks. The homogeneous prior is applied on all modules unless otherwise specified.
		\item \textit{Shared-Homo-Path, ACT-Homo-Path}: Models utilizing the fixed-depth and shared-weights paradigm (i.e. repeating one GN-block for multiple times) and the adaptive computation time algorithm~\cite{graves2016adaptive}, respectively. PathGNN and the homogeneous prior are all applied.
		\item \textit{Iter-Path}: Same as Iter-Homo-Path except that no homogeneous prior is applied.
		\item \textit{Iter-GAT}: Models utilizing our iterative module and adopting GAT as GN-blocks. No homogeneous prior is applied.
	\end{itemize}
	
	For most problems, max pooling is utilized as the readout function and we use only one IterGNN to build the core graph neural networks. The homogeneous prior is applied to all modules. However, for component counting, sum pooling is utilized as the readout function and two IterGNNs are stacked sequentially, since two iteration loops are usually required for component-counting algorithms (one for component assignment and one for counting). We utilize the node-wise IterGNN to support unconnected graphs as introduced in Section~\ref{sec:iterative-node-wise}. The homogeneous prior is only applied to the GNN modules but not the embedding module and count prediction module, because the problem is not homogeneous. Random initialization of node attributes is also applied to improve GNNs' representation powers as analyzed in Section~\ref{sec:random-init}.

	\subsection{Training Details}
	\label{sec:training-details}
	
	All models are trained with the same set of  hyper-parameters: the learning rate is 0.001, 
	and the batch size is 32. We use Adam as the optimizer. The hidden neuron number is 64. For models using the iterative module or the ACT algorithm, we train the networks with 30 maximal iterations and test them with no additional constraints. 
	For the fixed-depth shared-weights paradigm, we train the networks with 30 iterations and test them with 1000 iterations (maximum node numbers in the datasets). 
	The only two tunable hyper-parameter within our proposals are the epoch number$=$20, 40, $\cdots$, 200 and the degree of flexibilities of PathGNN, each corresponding to one variation of the PathGNN layer as described in Section \ref{appsec:pathgnn}. Another hyper-parameter within the ACT algorithm~\cite{graves2016adaptive} is $\tau = 0, 0.1, 0.01, 0.001$.
	We utilize the validation dataset to select the best hyper-parameter and report its performance on the test datasets.

	\section{Experimental results}
	\label{sec:exp-results}
	
	We present experimental results that are omitted in the main body due to the space limitation in Section~\ref{sec:exp-result-gen-size}. We then analyze the iteration numbers learned by ACT and our models in Section~\ref{sec:exp-result-iter-num}. In Section~\ref{sec:exp-results-pacman}, we present the generalization performances of IterGNN, GCN and PointNet for the symbolic Pacman task in environments with different number of dots and of walls.
	
	Other than those generalization performance w.r.t. scales, we evaluate the standard generalizability of our models on five graph classification benchmarks. As shown in Section~\ref{sec:exp-graph-classification}, our Iter-GIN achieves competitive performance to the state-of-art GIN~\cite{xu2018how}. We also verifies the claim in Section~\ref{sec:random-init} stating that regular graphs, which can not be distinguished by WL-test, can be distinguished by IterGNNs plus random initialization of node attributes, in Section~\ref{sec:exp-random-init}.
	
	\subsection{Solving graph theory problems}
	
	\subsubsection{Generalize w.r.t. graph sizes}
	\label{sec:exp-result-gen-size}
	
	We provide the omitted generalization performance of models on the weighted shortest path problem with PL and KNN as generators in Table~\ref{apptab:summary}. Each of our proposals help improve the generalizability of models with respect to graph sizes and graph diameters. Our final Iter-Homo-Path model largely outperforms the other models regarding the generalizability w.r.t. scales. 
	
	\subsubsection{The iteration numbers learned by ACT and our iterative module}
	\label{sec:exp-result-iter-num}
	
	We first analyze the ACT~\cite{graves2016adaptive} algorithm and show that it is easy for ACT to learn small iteration numbers. We then explain why the minimum depth of GNNs for accurately predicting the lengths $l$ of the shortest path is $l/2$ to show that our Iter-Homo-Path model actually learns an optimal and interpretable stopping criterion. At last, we evaluate the benefits of the decaying mechanism of our iterative module as described in Section~\ref{sec:iterative-decay}.
	
\begin{figure}[]
		\centering
		\includegraphics[width=0.6\textwidth]{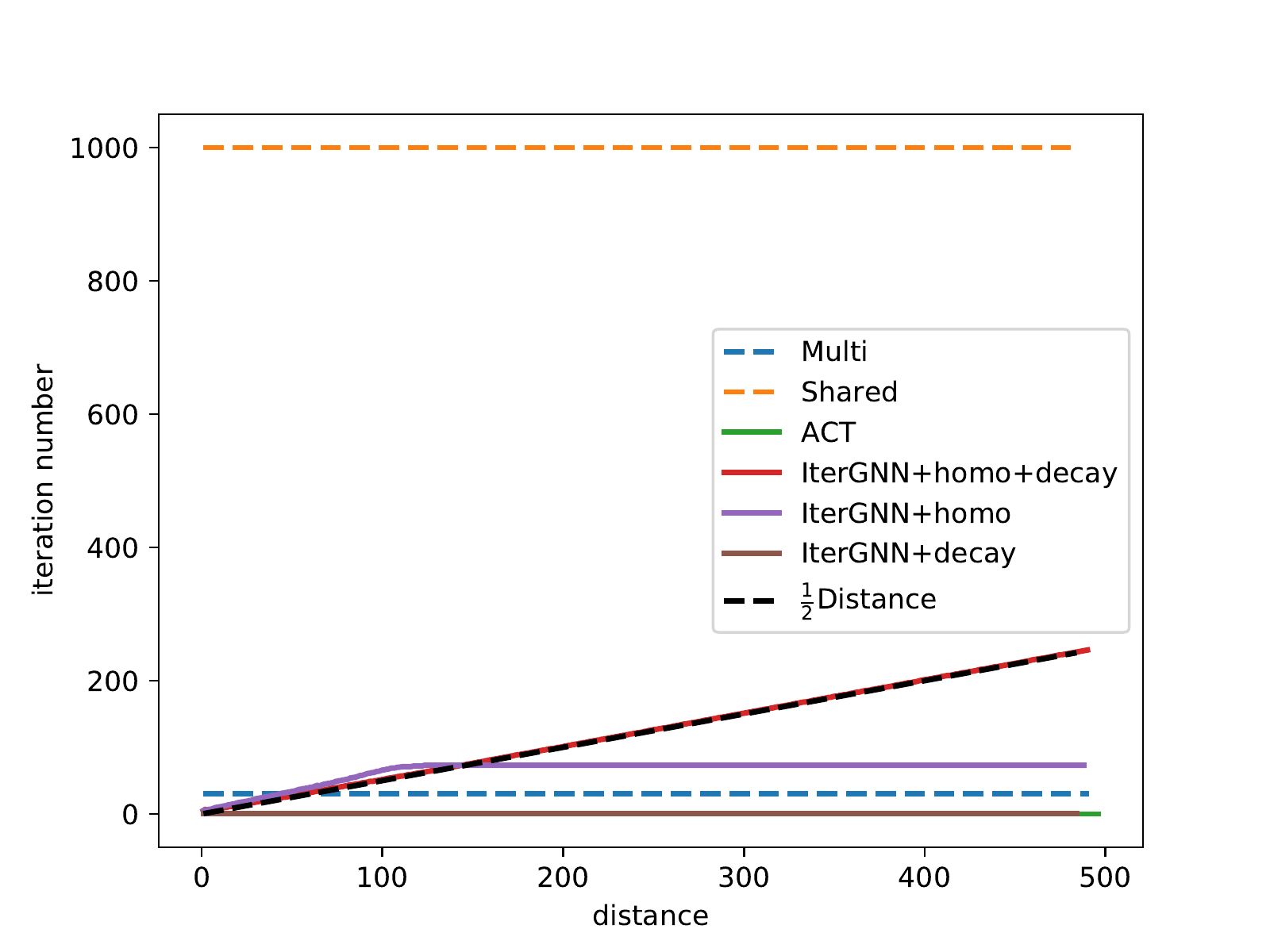}
		\caption{
			The iteration numbers of GNN layers w.r.t. the distances from the source node to the target node for the unweighted shortest path problem. All of them utilize Homo-Path as the backbone and change the paradigm to compose GN-blocks, except for the "IterGNN+decay" model. Multi denotes the simplest paradigm that stacks GN-blocks sequentially. The iteration numbers for the ACT algorithm and for the IterGNN models are all adaptive to the inputs and the stopping criterions are leanred during training. Models are trained on graphs of sizes $[4,34)$ while tested on graphs of diameters 500. The theoretical lower bound of iteration numbers for accurate prediction, i.e. distance/2, is also plotted.}
		\label{fig:layer_num}
	\end{figure}

	\paragraph{The iteration numbers learned by ACT are usually small.} As shown in Figure~\ref{fig:layer_num}, the \mbox{ACT-Homo-Path} model learns much smaller iteration numbers than other models. It is because of the formulations of the ACT algorithm. In detail, the ACT algorithm considers a different random process from our iterative module while designing the stopping criterion. The expected output has the form as $\sum_{i=1}^{k-1} c^ih^i+(1-\sum_{i=1}^{k-1}c^i)h^k$. The notations are the same as our iterative module described in the main body. And the stopping criterion is $\sum_{i=1}^{k}c^i > 1$. Compared with the expected output $\sum_{j=1}^{k}c^jh^j\prod_{i=1}^{j-1}(1-c^i)$ and the stopping criterion $\prod_{i=1}^{k}(1-c^i) < \epsilon$ of our iterative module, it is generally easier for the stopping criterion of ACT to be satisfied since $1-\sum_{i=1}^{k}c^i < \prod_{i=1}^{k}(1-c^i)$ if $k>1$ and $0<c^i<1$ for all $i<=k$. In the paper that proposes the ACT algorithm~\cite{graves2016adaptive}, the author also states similar intuitions that the formulation in our iterative module will not stop in a few steps. In fact, as far as we know, all works~\cite{graves2016adaptive, figurnov2017spatially, eyzaguirre2020differentiable} that utilize the ACT algorithm adopt a noisy regularization term to encourage fewer iteration numbers. They have distinct motivations and objectives from our work. For example, most of them are designed for more efficiencies by fewer iterations~\cite{figurnov2017spatially, eyzaguirre2020differentiable}. On the other hand, our iterative module is designed to improve the generalizability of GNNs with respect to scales by generalizing to much larger iterations.
	
	\paragraph{The minimum depth of GNNs to accurately solve the unweighted shortest path problem.} The minimum depth of GNNs to accurately predict the shortest path whose length is $l$ is $l/2$. 
	The reason is that, due to the message-passing nature of GN-blocks, $l$-layer GNNs can at most summarize information of the shortest paths whose lengths are smaller than $2l+1$. Therefore, if models can't iterate for distance/2 times, they can't collect enough information for making an accurate prediction of the shortest path lengths, but can only guess based on the graph's global properties (e.g. the number of nodes and edges) instead. As illustrated in Figure \ref{fig:layer_num}, our model, Iter+homo+decay, i.e. the Iter-Homo-Path model, learns the optimal stopping criterion, whose iteration numbers are equal to half of the shortest path length. In other words, it achieves the theoretical lower bound of the iteration numbers for accurate predictions given the message-passing nature of GN-blocks.

	\begin{table}
		\caption{
			Generalization performance on graph algorithm learning and graph-related reasoning. 
			Models are trained on graphs of sizes within $[4,34)$ and are tested on graph of larger sizes such as 100 (for the shortest path problem and the TSP problem) and 500 (for the component counting problem).
			The metric for the shortest path problem and the TSP problem is the relative loss. The metric for the component counting problem is the accuracy.
		}
		\label{apptab:summary}
		\centering
		\begin{tabular}{|c||cccc|cc|c|}
			\hline
			& \multicolumn{4}{c|}{Shortest Path - weighted} & \multicolumn{2}{c|}{Component Cnt.} &  \multicolumn{1}{c|}{TSP}\\
			Models & ER & PL & KNN & Lob & ER & Lob & 2D
			\\\hline
			GCN~\cite{kipf2017semi} & 0.1937 & 0.202 & 0.44 & 0.44& 0.0\% & 0.0\%& 0.52\\
			GAT~\cite{velickovic2018graph} & 0.1731 & 0.127 & 0.26 & 0.28 & 24.4 \% & 0.0\%& 0.18 \\
			Path (ours) & 0.0014 & 0.084 & 0.16 & 0.29 & 82.3\% & 77.2\%& 0.16\\
			Homo-Path (ours) & \textbf{0.0008} & 0.015 & 0.07 & 0.27 & \textbf{91.9\%} & 83.9\%& 0.14 \\ 
			Iter-Homo-Path (ours) & \textbf{0.0007} & \textbf{0.003} & \textbf{0.03} & \textbf{0.02} & 86.6\%& \textbf{95.4\%}& \textbf{0.11} \\ \hline
		\end{tabular}
	\end{table}

	\textbf{The benefits of the decaying mechanism of our iterative module.} 
	Let's compare the performance of IterGNN+homo+decay and IterGNN+homo in Figure~\ref{fig:layer_num}, to verify the benefits of the decaying mechanism. Although IterGNN+homo is still able to generalize to the number of iterations as large as 100, it can not generalize to much larger iteration numbers such as 200 or 2000 without the decaying mechanism. Models with fewer iteration numbers than the lower bound, distance$/2$, theoretically lack powers for accurately predicting the shortest paths of lengths larger than $l$. The success rate of IterGNN+homo is 67.5\%, which is much smaller than the success rate of IterGNN+homo+decay 100\%, for predicting the shortest path on lobster graphs of size $500$. The worse performance of IterGNN+homo than IterGNN+homo+decay suggests the effectiveness of our decaying mechanism for improving the generalizability of models with respect to graph scales.

	\subsection{Symbolic Pacman}
	\label{sec:exp-results-pacman}
	
	The experimental setups are presented in Section~\ref{sec:exp-setup-pacman}. Note that, unlike the original Atari PacMan environment, 
	our environment is more challenging because we randomly sample the layout of maps for each episode and we test models in environments with different numbers of dots and walls.
	The agent can not just remember one policy to get successful but needs to
	learn to do planning according to the current observation.
	
	Table~\ref{tab:pacman-itergnn}, Table~\ref{tab:pacman-gcn}, and Table~\ref{tab:pacman-pointnet} show the performance of IterGNN, GCN~\cite{kipf2017semi} and PointNet~\cite{qi2017pointnet}, respectively, in environments with different number of walls and dots. 
	Our IterGNN demonstrates remarkable generalizability among different environment settings, as stated in Table~\ref{tab:pacman-itergnn}. It successfully transfers policies to environments with different number of dots and different number of walls. IterGNN performs much better than the GCN and PointNet baselines, demonstrating that our proposals improve the generalizability of models. GCN performs the worst probably because of the unsuitable strong inductive bias encoded by the normed-mean aggregation module.
	
	\begin{table}
		\centering
		\caption{Generalization performance of IterGNN for symbolic Pacman. Metric is the success rate of eating dots.
			Models are trained in environments with 10 dots and 8 walls
			and are tested in environments with different number of walls and dots. }
		\label{tab:pacman-itergnn}
		\begin{tabular}{|c|ccccc|}\hline
			\diagbox{\#wall}{\#dots}& 1 & 5 & 10 & 15 & 20  \\\hline
			0 & 1.00 & 1.00 & 0.99 & 0.99 & 1.00 \\
			3 & 0.95 & 1.00 & 0.98 & 0.94 & 0.98 \\
			6 & 0.90 & 0.95 & 0.94  & 0.97 & 0.98 \\
			9 & 0.80 & 0.92 & 0.95 & 0.95 & 0.93 \\
			12 & 0.60 & 0.96 & 0.97 & 0.98 & 0.93 \\
			15 & 0.75 & 0.92 & 0.94 & 0.95 & 0.97 \\\hline
		\end{tabular}
	\end{table}
	\begin{table}
		\centering
		\caption{Generalization performance of GCN for symbolic Pacman. Metric is the success rate of eating dots.
			Models are trained in environments with 10 dots and 8 walls
			and are tested in environments with different number of walls and dots. }
		\label{tab:pacman-gcn}
		\begin{tabular}{|c|ccccc|}\hline
			\diagbox{\#wall}{\#dots}& 1 & 5 & 10 & 15 & 20  \\\hline
			0 & 0.05 & 0.23 & 0.09  & 0.07  & 0.05  \\
			3 & 0.10  & 0.23 & 0.20  & 0.15  & 0.04  \\
			6 & 0.00  & 0.32  & 0.17   & 0.09  & 0.06  \\
			9 & 0.20  & 0.27  & 0.23  & 0.06  & 0.10  \\
			12 & 0.05  & 0.18  & 0.26  & 0.13  & 0.03  \\
			15 & 0.05  & 0.17  & 0.23  & 0.15  & 0.08  \\\hline
		\end{tabular}
	\end{table}\begin{table}
		\centering
		\caption{Generalization performance of PointNet for symbolic Pacman. Metric is the success rate of eating dots.
			Models are trained in environments with 10 dots and 8 walls
			and are tested in environments with different number of walls and dots. }
		\label{tab:pacman-pointnet}
		\begin{tabular}{|c|cccccc|}\hline
			\diagbox{\#wall}{\#dots}& 1 & 3 & 6 & 9 & 12 & 15  \\\hline
			2 & 0.82 & 0.58 & 0.39 & 0.32  & 0.34 & 0.29 \\
			4 & 0.72  & 0.48  & 0.31  & 0.25  & 0.24 & 0.22 \\
			6 & 0.71  & 0.31  & 0.36   & 0.24  & 0.19 & 0.14 \\
			8 & 0.60  & 0.36  & 0.21  & 0.20  & 0.20 & 0.28 \\
			10 & 0.50  & 0.33  & 0.33  & 0.29  & 0.16 & 0.15 \\\hline
		\end{tabular}
	\end{table}

	\subsection{Graph Classification}
	\label{sec:exp-graph-classification}
	
	At last, we evaluate models on standard graph classification benchmarks to show that our proposals do not hurt the standard generalizability of GNNs while improving their generalizability w.r.t. scales. More descriptions of the task are available in \cite{xu2018how}. The experimental setups are presented in Section~\ref{sec:exp-setup-graph-classification}.
	
	As stated in Table \ref{tab:benchmarks}, our model performs competitively with the previous state-of-art backbone, GIN, on all five benchmarks. It suggests that our iterative module is a safe choice for improving generalizability w.r.t. scales while still maintaining the performance for normal tasks. Note that, due to the shortage of time, little hyper-parameter search was conducted in our experiments. Default hyper-parameters such as learning rate equal to 0.001 and hidden size equal to 64 were adopted. Therefore, the performance of Iter-GIN is potentially better than those stated in Table \ref{tab:benchmarks}.
	
	\begin{table}[]
		\centering
		\begin{tabular}{|c|c|c|}\hline
			Dataset & GIN & Iter-GIN\\\hline
			IMDB-B & 75.1$\pm$5.1 & \textbf{75.7$\pm$4.2} \\\hline
			IMDB-M & \textbf{52.3$\pm$2.8} & 51.8$\pm$4.0 \\\hline
			MUTAG & 89.4$\pm$5.6 & \textbf{89.6$\pm$8.6} \\\hline
			PROTEINS & 76.2$\pm$2.8 & \textbf{76.3$\pm$3.4} \\\hline
			PTC  & \textbf{64.6$\pm$7.0} & 64.5$\pm$3.8 \\\hline
		\end{tabular}
		\caption{The performance of our iterative module on graph classification on five popular benchmarks. Iter-GIN is built by wrapping each GIN module in the previous state-of-art method~\cite{xu2018how} using our iterative module. Metric is the averaged accuracy and STD in 10-fold cross-validation.}
		\label{tab:benchmarks}
	\end{table}
	
	\subsection{Effectiveness of Random Initialization}
	\label{sec:exp-random-init}
	
	As discussed in Section \ref{sec:random-init}, we adopt random initialization to improve GNNs' representation powers especially for solving graph-related problems such as component counting and graph diameters. The argument is based on that non-isomorphic regular graphs with random initialized node attributes are distinguishable by GNNs, i.e. GNNs can distinguish sets of graphs as illustrated in Figure \ref{fig:random-init}(d). Although it has been proved theoretically in the bounded settings \cite{murphy2019relational, anonymous2020coloring}, we further verify its effectiveness in practice. 
	
	The task is then as simple as a binary classification problem to distinguish regular graphs as shown in Figure \ref{fig:random-init}(a) and Figure \ref{fig:random-init}(b). 10000 samples are generated for training with graph structures uniformly sampled from two regular graphs and the node attributes uniformly sampled between 0 and 1. One thousand samples are then generated randomly for test.  No validation set is needed as we do not perform hyper-parameter tuning. Experimental results show that our Iter-HomoPath model achieves 100\% accuracy for distinguishing both pairs of non-isomorphic regular graph.

	\section{Backgrounds - Graph Neural Networks}
	\label{sec:background}
	
	In this section, we briefly describe the graph neural networks (GNNs). We first present the GN blocks, which generalize many GNN layers, in Section~\ref{sec:gn-block} and then present the common practice of building GNN models for graph classification/regression in Section~\ref{sec:gnns}. The notations and terms are further utilized while describing PathGNN layers in Section~\ref{appsec:pathgnn} and while describing the models/baselines in our experiments in Section~\ref{sec:exp.models}.
	
	\subsection{Graph Network Blocks (GN blocks)}
	\label{sec:gn-block}
	
	We briefly describe a popular framework to build GNN layers, called Graph Network blocks (GN blocks)~\cite{battaglia2018relational}. It encompasses our PathGNN layers presented in Section~\ref{appsec:pathgnn} as well as many state-of-art GNN layers, such as GCN~\cite{kipf2017semi}, GAT~\cite{velickovic2018graph}, and GIN~\cite{xu2018how}. Readers are referred to~\cite{battaglia2018relational} for more details. Note that we adopt different notations from the main body to be consistent with~\cite{battaglia2018relational}. Also note that, even when the global attribute is utilized, the fixed-depth fixed-width GNNs still lose a significant portion of powers for solving many graph problems as proved in~\cite{Loukas2020What}. For example, the minimum depth of GNNs scales sub-linearly with the graph sizes for accurately verifying a set of edges as the shortest path or a s-t cut, given the message passing nature of GNNs.
	
	The input  graphs are defined as $G=(\textbf{u},V,E)$ with node attributes $V=\{\textbf{v}_i, i=1,2,\dots,N_v\}$, edge attributes $E=\{(\textbf{e}_k, s_k, r_k),  k=1,2,\dots,N_e\}$, and the global attribute $\textbf{u}$, where $s_k$ and $r_k$ denote the index of the sender and receiver nodes for edge $k$, $\textbf{u}$, $\textbf{v}_i$, and $\textbf{e}_k$ represent the global attribute vector, the attribute vector of node $i$, and the attribute vector of edge $k$, respectively.
	
	The GN block performs message passing using three update modules $\phi^e, \phi^v, \phi^u$ and three aggregation modules $\rho^{e\rightarrow v}, \rho^{e\rightarrow u}, \rho^{v\rightarrow u}$ as follows:  
	
	\begin{enumerate}
		\item Node $s_k$ sends messages $\textbf{e}_k'$ to the receiver node $r_k$, which are updated according to the related node attributes $\textbf{v}$, edge attributes $\textbf{e}$, and global attributes $\textbf{u}$. $$\textbf{e}_k'=\phi^e(\textbf{e}_k, \textbf{v}_{r_k}, \textbf{v}_{s_k}, \textbf{u}), k=1,2,\dots,N_e$$
		\item For each receiver node $i$, the sent messages are aggregated using the aggregation module. $$\bar{\textbf{e}}_i'=\rho^{e\rightarrow v}(\{\textbf{e}'_k | r_k=i\}$$
		\item The aggregated messages are then utilized for updating the node attribute together with the related node attributes $\textbf{v}_i$ and global attributes $\textbf{u}$.
		$$\textbf{v}_i' = \phi^v(\bar{\textbf{e}}_i', \textbf{v}_i, \textbf{u})$$ 
		\item The sent messages $\textbf{e}_k'$ can also be aggregated for updating the global attribute $\textbf{u}$ as $$\bar{\textbf{e}}'=\rho^{e\rightarrow u}(\{\textbf{e}'_k, k=1,2,\dots,N_e\}$$
		\item The node attribute $\textbf{v}_i'$ can be aggregated for updating the global attribute $\textbf{u}$ as well. $$\bar{\textbf{v}}'=\rho^{v\rightarrow u}(\{\textbf{v}'_i, i=1,2,\dots,N_v\}$$ 
		\item The global attribute $\textbf{u}$ are updated according to the aggregated messages $\bar{\textbf{e}}'$, aggregated node attributes $\bar{\textbf{v}}'$, and previous global attribute $\textbf{u}$. $$\textbf{u}' = \phi^u(\bar{\textbf{e}}', \bar{\textbf{v}}', \textbf{u})$$
	\end{enumerate} 
	$\phi^e$ is often referred as the message module, $\rho^{e\rightarrow v}$ as the aggregation module, $\phi^v$ as the update module, $\rho^{e\rightarrow u}$ and $\rho^{v\rightarrow u}$ as the readout modules, in many papers. The three update modules are simple vector-to-vector modules such as multi-layer perceptrons (MLPs). The three aggregation modules, on the other hand, should be permutation-invariant functions on sets such as max pooling and attentional pooling \cite{li2015gated, lee2019self}.

	\subsection{Composing GN blocks for graph classification / regression}
	\label{sec:gnns}
	
	We follow the common practice~\cite{battaglia2018relational,zhang2018end,xu2018how} in the field of supervised graph classification while building models and baselines in our experiments. Typically, the models are built by sequentially stacking the node-wise embedding module, the core GNN module, the readout function, and the task-specific prediction module. The node-wise embedding module corresponds to GN blocks that are only composed of function $\phi^v$ for updating node attributes. More intuitively, it applies the same MLP module to update all node attribute vectors. The core GNN module performs message passing to update all attributes of graphs. The readout function corresponds to GN blocks that only consist of the readout modules such as $\rho^{e\rightarrow u}$ and $\rho^{v\rightarrow u}$ to summarize information of the whole graph into a fixed-dimensional vector $\textbf{u}$. The task-specific prediction module then utilizes the global attribute vector $\textbf{u}$ to perform the final prediction, such as predicting the number of connected components or the Q-values within the symbolic Pacman environment.
	
	We need to specify three properties while designing the core GNN modules: (1) the internal structure of GN blocks; (2) the composition of GN blocks; and (3) the prior that encodes the properties of the solutions of the problem. 
	\begin{itemize}
		\item The internal structure of GN blocks defines the logic about how to perform one step of message passing. It is usually specified by selecting or designing the GNN layers. 
		\item The composition of GN blocks defines the computational flow among GN-blocks. For example, the simplest paradigm is to apply multiple GN blocks sequentially. Our iterative module introduces the iterative architecture into GNNs. It applies the same GN block for multiple times. The iteration number is adaptively determined by our iterative module. 
		\item The prior is usually specified by adopting the regularization terms. For example, regularizing the L2 norm of weights can encode the prior that GNNs representing solutions of the problem usually have weights of small magnitudes. Regularizing the L1 norm of weights can encode prior about sparsity. We can utilize the HomoMLP and HomoGNN as described in the main body to encode the homogeneous prior that the solutions of most classical graph problems are homogeneous functions.
	\end{itemize}


\end{document}